\documentclass[conference]{IEEEtran}
\usepackage{multirow}
\usepackage{amsmath}
\usepackage{amssymb}
\usepackage{xcolor}
\usepackage{tikz}
\usepackage{makecell}
\usepackage{lscape}
\usepackage{pifont}
\newcommand{\cmark}{\ding{51}}%
\newcommand{\xmark}{\ding{55}}%
\usepackage[graphicx]{realboxes}
\usepackage{adjustbox}

\usepackage{hyperref}

\usepackage{array}
\newcolumntype{P}[1]{>{\centering\arraybackslash}p{#1}}
\newcolumntype{C}[1]{>{\centering\arraybackslash}p{#1}}

\ifCLASSINFOpdf
\else
\fi
\usepackage{comment}
\usepackage{amsmath}
\usepackage{amsfonts}
\usepackage{graphicx}
\usepackage{caption}
\usepackage{marvosym}
\begin{document}
%

\title{Face Deepfakes - A Comprehensive Review}


\author{\IEEEauthorblockN{Tharindu Fernando\IEEEauthorrefmark{1}, Darshana Priyasad\IEEEauthorrefmark{1},
Sridha Sridharan\IEEEauthorrefmark{1},  Arun Ross\IEEEauthorrefmark{2}, and
Clinton Fookes\IEEEauthorrefmark{1}.}\\

\IEEEauthorblockA{\IEEEauthorrefmark{1}Signal Processing, Artificial Intelligence \& Vision Technologies, Queensland University of Technology, Australia.
}\\

\IEEEauthorblockA{\IEEEauthorrefmark{2}Michigan State University, Department of Computer Science and Engineering, United States.}\\
\thanks{Corresponding author: T. Fernando (email: t.warnakulasuriya@qut.edu.au).}}

\maketitle

\begin{abstract}
In recent years, remarkable advancements in deepfake generation technology have led to unprecedented leaps in its realism and capabilities. Despite these advances, we observe a notable lack of structured and deep analysis deepfake technology. The principal aim of this survey is to contribute a
thorough theoretical analysis of state-of-the-art face deepfake generation and detection methods. Furthermore, we provide a coherent and systematic evaluation of the implications of deepfakes on face biometric recognition approaches. In addition, we outline key applications of face deepfake technology, elucidating both positive and negative applications of the technology, provide a detailed discussion regarding the gaps in existing research, and propose key research directions for further investigation.
\end{abstract}


%
\IEEEpeerreviewmaketitle

\section{Introduction}
On the 1st of June, 2019 artists Bill Posters and Daniel Howe released a deepfake video \footnote{https://billposters.ch/the-zuckerberg-deepfake-heard-around-the-world/} that featured Mark Zuckerberg delivering a speech about the power of Facebook and its control over user data. This synthesised video was generated to raise awareness regarding deepfakes and their potential implication. Soon after its release, this video made it into the major global news platforms, including, the New York Times, ABC News, and BBC, and sparked discussions about the ethical and social implications of deepfake technology.


It has been five years since the release of the Mark Zuckerberg deepfake video and during this period there have been significant technological advancements in deepfake generation technology On the other hand,  advances in deepfake detection have lagged behind and there has been very little response, over this 5-year duration from regulators and educators to educate and safeguard society from the malicious use of deepfake technology. The principal aim of this study is to provide up-to-date algorithmic insights regarding face deepfake generation and detection processes.  Our systematic analysis is not only beneficial to researchers and machine learning practitioners but is also of pertinent interest to the general public and policymakers and we expect this review to raise awareness among these groups regarding both positive and negative implications of deepfakes.


\subsection{What are deepfakes?}
The term deepfakes has been used to refer to any synthetic media that has been generated using deep learning techniques, including the media generated using generative AI technology. However, a clear distinction exists between generative AI and deepfakes generation when considering their purpose. The purpose of generative AI is mainly to generate synthetic content by analysing the patterns of the real-data while deepfakes are primarily designed to generate realistic-looking content with the main aim to fool the users of that media into thinking it is real. Deepfakes first came into focus in 2017 when a Reddit user released fake celebrity pornographic videos generated using the deepfake technology.


\subsection{Why is it important to be educated about deepfakes?}
The 2024 global risk report \cite{wef2024} of the World Economic Forum identified that misinformation and disinformation are the biggest short-term risks to the world economy. A Boston University article \cite{Weinstein2021} suggested that international and domestic disinformation campaigns targeting Americans are America’s greatest national security threat, which is a greater danger than the nuclear capabilities of Russia, China, and North Korea. Deepfake technology plays a major role in generating highly convincing but entirely fabricated faces, voices, and text making it difficult for people to discern truth from fiction. Therefore, understanding the process of creation, the impact of deepfakes, and a study of the existing detection technologies in terms of their effectiveness is crucial in today's digital age due to the rapidly growing presence of deepfakes. 

While deepfakes can span various mediums of communication including text, audio, images, and video, in this article we limit our discussion to images, and video-based deepfakes of human faces as they are the most powerful mediums that the perpetrators can leverage to spread misinformation, manipulate public opinion, and deceive individuals. Understanding deepfakes would help organisations to analyse the potential vulnerabilities of the systems they utilise and apply mechanisms to mitigate these threats. Moreover, by staying informed about deepfake technology society can be better prepared for its negative implications and be vigilant by critically evaluating the media they consume. Furthermore, a better understanding of both the positive and negative impacts of deepfake technology is essential for recognising the potential for misuse of this technology, promoting responsible practices, and fostering ethical development and use of this technology. Our review paper takes an important step toward this by providing a comprehensive and systematic analysis of existing literature on face deepfake generation and detection with a special focus on their implications in biometric recognition.

\subsection{How is our study different from existing surveys?}

While there exist several surveys that discuss the generation and detection of face deepfakes, we observe a lack of studies that provide an in-depth algorithmic overview including a discussion regarding training paradigms, the loss functions, and evaluation metrics. In Tab. \ref{tab:comparison_to_other_surveys} we summarise the main topics that our paper covers and compare it with the coverage of existing works.

\begin{table*}[htbp]
\caption{Comparison of our survey to other related studies. Note: $*$ indicates a comprehensive discussion, and $+$ indicates just an overview.}
 \resizebox{\textwidth}{!}{%
\begin{tabular}{|c|l|c|c|c|c|c|c|}
\hline
Paper  &Year& Face Deepfakes Generation &Face Deepfakes Detection&Algorithimic Discussion& Applications of Deepfakes & Impact on Biometric Recongition  &Future Research Directions\\ \hline

    Dagar et. al \cite{dagar2022literature}  &2022&  \cmark ($*$)&\cmark($*$)&\cmark ($+$)&                          \cmark ($+$)&    \xmark                               &\cmark ($+$)\\ \hline
   Nguyen \cite{nguyen2022deep}    &2022&  \cmark($*$)&\cmark($*$)&\cmark($+$)&     \xmark                      & \xmark                                 &\cmark ($+$)\\ \hline
  Waseem et. al \cite{waseem2023deepfake} &2023&  \cmark ($*$)&\cmark ($*$)&\cmark($+$)&  \xmark                          &      \xmark                              &\cmark($*$)\\ \hline
 Masood et. al \cite{masood2023deepfakes}     &2023&  \cmark($*$)&\cmark($*$)&\xmark  &          \xmark                 &       \xmark                           &\cmark ($+$)\\ \hline
 Mubarak et. al \cite{mubarak2023survey}     &2023&  \xmark  &\cmark($*$)&\xmark  &          \xmark                 &       \cmark ($+$)&\xmark                           \\ \hline
  
   Patel et. al \cite{patel2023deepfake}     &2023&  \cmark($*$)&\cmark($*$)&\cmark ($*$) &          \xmark                 &       \xmark                           &\xmark                           \\ \hline
    Wang et. al \cite{wang2024deepfake}     &2024&  \xmark  &\cmark($*$)&\cmark ($+$)&          \xmark                 &       \xmark                           &\xmark                           \\ \hline
    Heidari et. al \cite{heidari2024deepfake}     &2024&  \xmark  &\cmark($*$)&\xmark&          \xmark                 &       \xmark                           &\cmark ($+$)\\ \hline
   Passos et. al \cite{passos2024review}     &2024& \xmark  & \cmark($*$)&\cmark ($+$)&          \xmark                 &       \xmark                           &\cmark ($+$)\\ \hline
   Sharma et. al \cite{sharma2024systematic}     &2024&  \xmark & \cmark($*$)& \cmark ($*$) &  \cmark ($+$)        &       \xmark                           &\cmark ($*$)\\\hline
    Ours    &2025&  \cmark($*$) & \cmark($*$)& \cmark ($*$) &  \cmark($*$)       &      \cmark($*$)                      &\cmark($*$)\\\hline
 
\end{tabular}}
\label{tab:comparison_to_other_surveys}
\end{table*}

\subsection{Organisation}
\begin{figure*}[htbp]
    \centering
    \includegraphics[width=\linewidth]{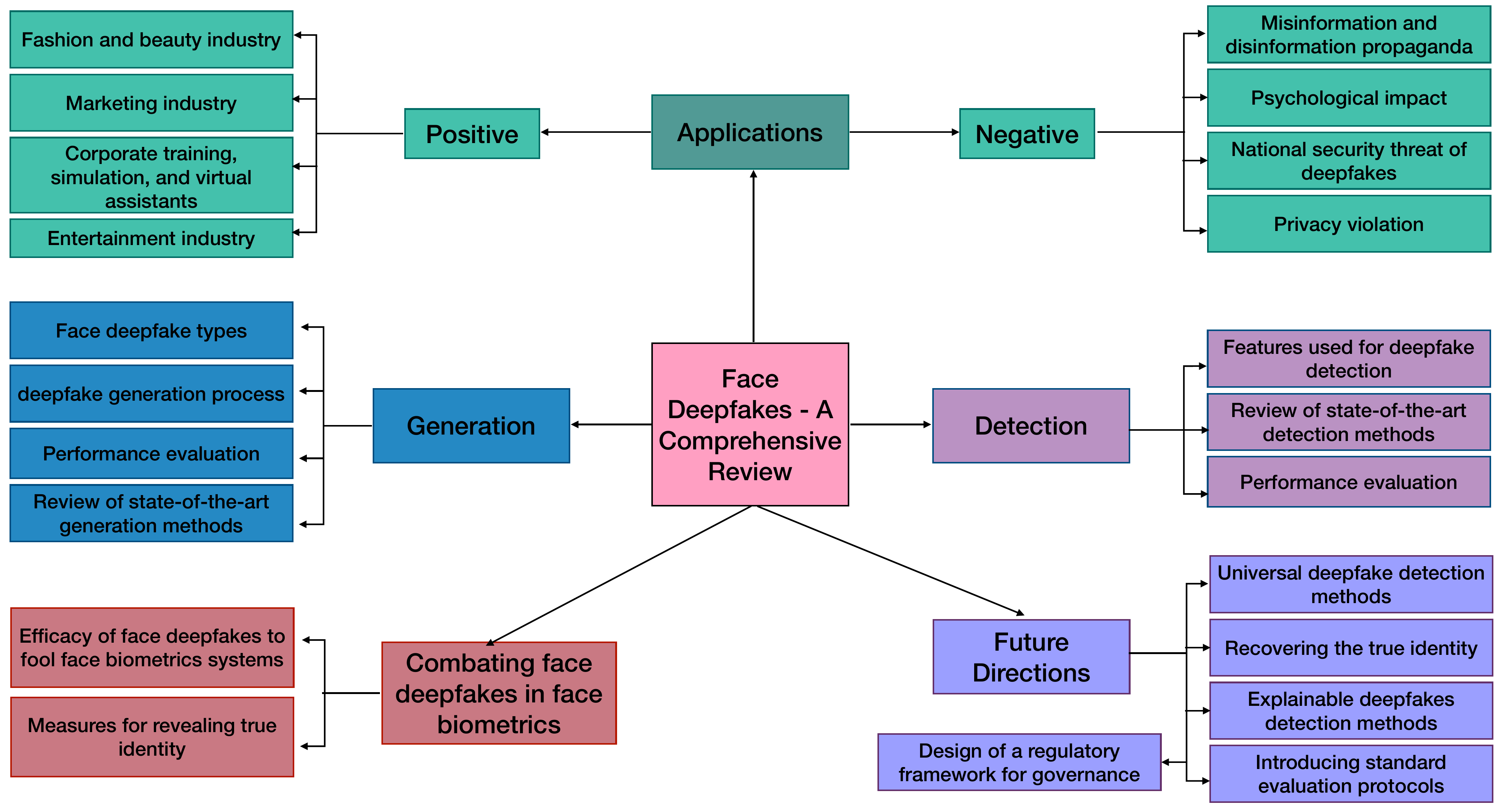}
    \caption{The Organization of this Survey}
    \label{fig:organisation}
\end{figure*}

The rest of our paper has the structure illustrated in Fig. \ref{fig:organisation}. Sec.  \ref{sec:face_deepfakes} discusses face deepfakes, with separates subsection devoted for both the generation and detection technologies. Under generation of face deepfakes, we first introduce different deepfake types, provide an overview of the deepfake generation process, illustrate popular evaluation metrics used to evaluate the quality of the generated deepfakes, and review the state-of-the-art approaches for the generation of deepfakes. Under the detection subsection of face deepfakes, we first analyse the features that have been leveraged to detect those deepfakes. Then we conduct a review of state-of-the-art methodologies for deepfake detection and introduce the evaluation metrics that have been introduced to evaluate the detection performance. Moreover, we quantitatively analyse the efficacy of the generated deepfake media to fool the state-of-the-art biometric recognition models. In addition, we also discuss the methodologies that have been introduced to uncover the true identity from the manipulated media. Furthermore, Sec. \ref{sec:applications} of our paper dicusses the applications of deepfakes, including both positive and negative applications. Sec. \ref{sec:future_research}  illustrates some of the challenges and limitations of existing literature on deepfakes frameworks, and provides future directions to pursue. Sec. \ref{sec:conclusions} contains concluding remarks.

\section{Face deepfakes}\label{sec:face_deepfakes} 
\subsection{Generation}

\begin{figure*}[htbp]
    \centering
    \includegraphics[width=\textwidth]{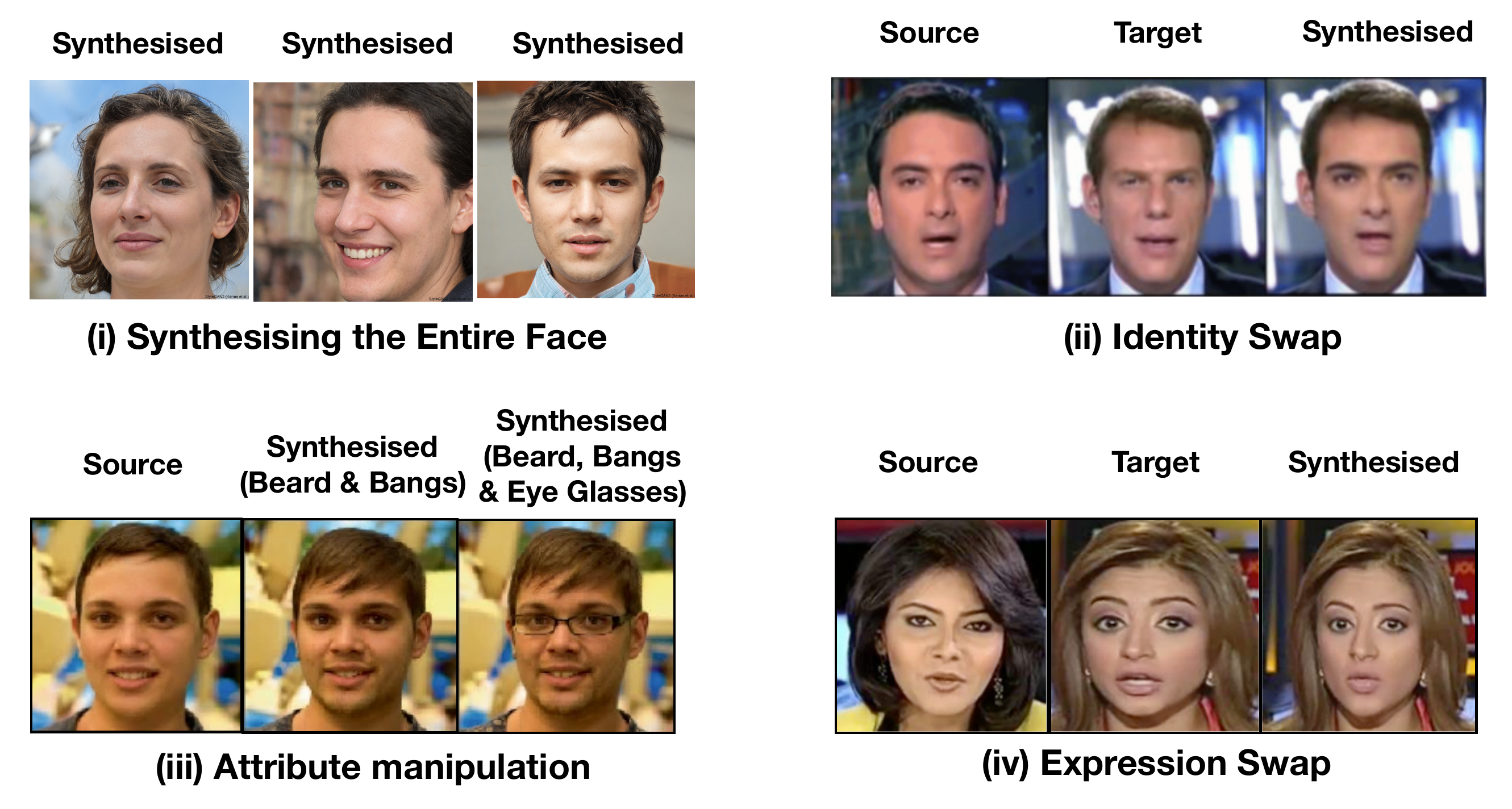}
    \caption{Illustration of different face generation techniques within deepfakes. Sub-figures (i)-(iv) have been sourced from {\tiny\protect\footnotemark[2]}, {\tiny\protect\footnotemark[3]}, {\tiny\protect\footnotemark[4]}, and {\tiny\protect\footnotemark[5]}, respectively.}
    \label{fig:face_generation_types}
\end{figure*}

Face manipulation techniques within deepfakes can be broadly categorised into four groups based on the level of manipulation. (i) synthesising the entire face: creating a non-existent face using generative AI technology, (ii) identity swap: replacing the face of one person in a video with another one, (iii) attribute manipulation: modifying some facial attributes such as eyeglasses, hair color, etc., and (iv) expression swap: modifying facial expressions in an image or video. Fig. \ref{fig:face_generation_types} visually illustrates differences between these generation types. 



\footnotetext[2]{https://thispersondoesnotexist.com/}
\footnotetext[3]{https://www.mdpi.com/2076-3417/13/11/6711}
\footnotetext[4]{https://link.springer.com/chapter/10.1007/978-3-031-19778-9\_41}
\footnotetext[5]{https://ieeexplore.ieee.org/abstract/document/10285057/}

Deep learning-based face swap models are capable of replacing one person's face in an image or video with another person's face, maintaining the overall structure and movement of the original face \cite{mirsky2021creation}. As such they can seamlessly attain identity swap face manipulation. The word reenactment implies acting out a past event or bringing something to life. Similarly, facial reenactment refers to bringing the source image to life by modifying it based on the movement of the head, lips, and facial expression in the target video (also called as driving video). Therefore, facial reenactment methods fall under the expression swap category and are not intended to directly alter a person's identity in a video. However, the recent advances that facial reenactment technology attained have enabled it to achieve far-reaching flexibility, that extends beyond simple expression manipulations and is of significant concern. As such review both face swap and face reenactment literature in this section.

Despite its negative implications such as misinformation and fake news, impersonation, identity theft, deepfake pornography, face swap technology has its faithful applications such as therapeutic and psychological applications and entertainment. For instance, face swap technology is already being used in exposure therapy and empathy-building exercises \cite{yang2022can}. Furthermore, off-the-shelf face swap apps such as Deepswap \footnote{https://deepgram.com/ai-apps/deepswap} and Faceswapper \footnote{https://faceswapper.ai/} are readily being used in social media for creating humorous or creative content by swapping faces with celebrities or other popular characters. As such research on improving the quality and realism of the face swap content has been at the forefront of machine learning research.

Along the same lines face reenactment or in other words talking face generation becoming increasingly popular with each passing day as it opens up a multitude of novel applications in this digital age. Video conferencing by animating a well-dressed image of ourselves without the need to transmit a live video stream \cite{deepfakevideocall} or commercials in which human actors are replaced by the faces generated by deepfakes \cite{lomnitz2020multimodal} which would have seemed a fantasy a few years a go has now become a reality thanks to advances of face reenactment technology.

\subsubsection{Face deepfake types}

The face swap models can be generally categorised based on the algorithm that they utlise in the face swap process. For instance, landmark-based methods where facial landmarks are first detected in both the source and target faces to guide the swapping process, auto-encoder-based end-to-end learning models, Generative Adversarial Networks (GANs) based approaches, etc. The technical details of these approaches are discussed in detail in Sec. \ref{sec:DeepfakeGenerationProcess}.

Existing works on face reenactment can be broadly categorised based on the type of modality used to drive the reenactment into three groups: (i) video-driven face reenactment methods, (ii) audio-driven face reenactment methods, and (iii) text-driven face reenactment methods. 

Video-driven methods where information from the driving video is used to extract the features required to reenact a source image are considered the most powerful compared to audio-driven and text-driven methods. Despite the fact that the identity of the person in the source image and the driving video could be different, motion, expression, and other facial features could be extracted from the driving video, allowing the video-driven method to extract a richer feature. However, audio and text-driven methods offer more practicality in real-world applications such as video conferencing, film production, or augmented reality where obtaining a driving video is impractical. However, obtaining driving audio or text is more feasible \cite{agarwal2023audio}.

\subsubsection{deepfake generation process}\label{sec:DeepfakeGenerationProcess}
In this section we provide an overview of the techniques utilised in literature for face swapping and face reenactment.

\begin{figure*}
    \centering
    \includegraphics[width=\textwidth]{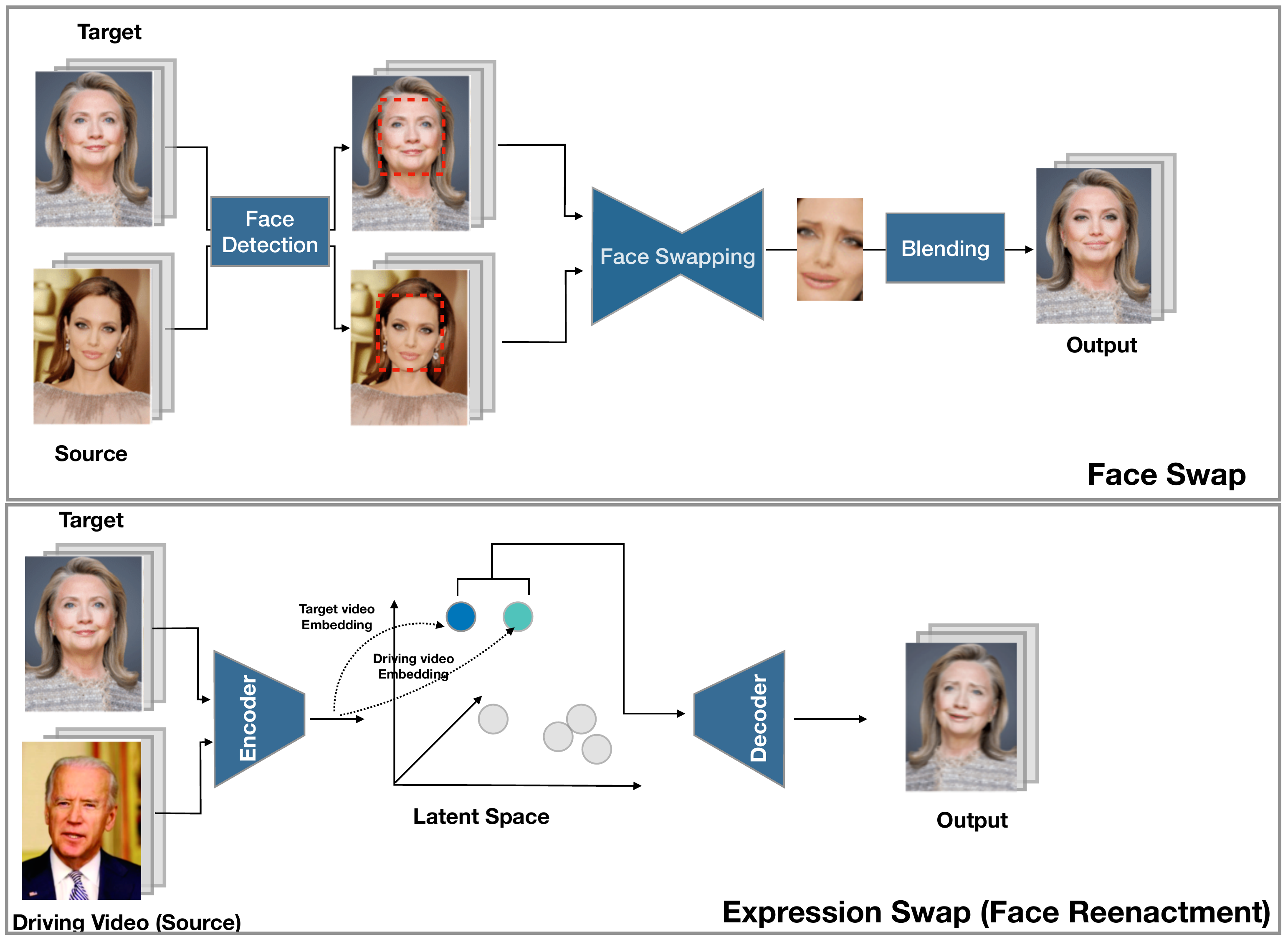}
    \caption{A comparison between face swapping and face reenactment processes}
    \label{fig:DeepfakeGenerationProcess}
\end{figure*}

Fig. \ref{fig:DeepfakeGenerationProcess} provides a comparison between the processes involved in face swapping and face reenactment. In general, face swap algorithms have three main steps \cite{waseem2023deepfake}. First, they detect the faces in both the source and target video, and the main attributes of the face in the target video such as nose, mouth, and eyes are replaced by the corresponding features of the source face. The next step involves the blending of the manipulated attributes to match the target video's colour and lighting. 

In contrast, in the face reenactment process, both the source image and the driving video are encoded into a latent space. Lower dimensional motion representations that capture head pose, expression, etc. are extracted from the latent space of the target video, and the identity information from the latent space of the source image. The decoder leverages this information and animates the source image using a driving video’s motion while preserving the source identity \cite{agarwal2023audio}.  
 
Autoencoders, Generative Adversarial Networks, Latent Space Decomposition approaches, and Diffusion Models are among the most popular techniques utilised in literature for face swapping and face reenactment, and the rest of the section provides an overview of these techniques and how they have been leveraged in the deepfake generation process

\textbf{Autoencoders: } Based on an encoder-decoder architecture, autoencoders are driven by the principle of learning a compressed latent representation of the input data that captures the salient attributes. The encoder encodes the input data into this learned latent space and the decoder should be able to use these latent embeddings and reconstruct the input without any information loss. The training process is guided by a loss function that minimises the reconstruction error and among the loss functions leveraged in the literature Mean Squared Error (MSE) Loss and Kullback-Leibler (KL) Divergence Loss are popular.

MSE loss can be written as,

\begin{equation}
    L_{MSE}(x, \hat{x}) = \frac{1}{N} \sum_{i=1}^{N} (x_i - \hat{x}_i)^2,
\end{equation}
where $x$ denotes input data, $\hat{x}$ denotes the reconstruction of the input data using the latent embeddings and $N$ is the number of samples in input data.

KL divergence loss measures the discrepancy between the distribution of the encoded latent representations and  predefined prior distribution and can be calculated as,

\begin{equation}
L_{KL}(q(z|x) \| p(z)) = -\frac{1}{2} \sum_{i=1}^{N} \left(1 + \log(\sigma_i^2) - \mu_i^2 - \sigma_i^2\right),
\end{equation}
$q(z|x)$ is the distribution of the encoded latent representations, $p(z)$ denotes predefined prior distribution, and $\mu_i$, and $\sigma_i$ are the mean and standard deviation of the latent representation of the $i$th input, respectively.

Numerous architectures \cite{perov2020deepfacelab, dfaker, tewari2017mofa} have been proposed that make use of the autoencoder technique for face swapping and face reenactment. While they have intricate differences they are based on the principle of learning a latent space that captures salient facial characteristics and identity information and decoding that information onto the target video. 

\textbf{Generative Adversarial Network (GAN)}: Inspired by the recent success of GANs for generating photo-realistic synthetic content, numerous works have leveraged GANs for generating face deepfakes. GANs also operate under the same principle of encoder-decoder architecture, however, additional supervision is provided via a discriminator, $D$. Specifically, the encoder of the generator $G$ maps the source data $x$ into a latent embedding $\phi$ i.e. $x \rightarrow \phi$, and the decoder portion of $G$ utilises this latent embedding for decoding the target representation $\hat{y}$ i.e $\phi \rightarrow y$. To augment the learning of mapping $x \rightarrow \phi \rightarrow \hat{y}$ an adversarial objective is proposed where the goal is to make the generated synthetic content look realistic such that the discriminator cannot differentiate between real and generated content. This objective can be written as,

\begin{equation}
\begin{split}
L_{GAN} & = \min_G \max_D \mathbb{E}_{x \sim p_{{data}}(x)} [\log D(x)] \\
& + \mathbb{E}_{z \sim p_z(z)} [\log (1 - D(G(z)))]
\end{split}
\end{equation}

where $y$ is the ground truth target and $z$ is random noise. 

In addition to the adversarial objective, several works have utilised additional loss terms such as L1 reconstruction loss between $\hat{y}$ and $y$ \cite{wu2018reenactgan} or perceptual loss which calculates the differences between high-level feature representations extracted from pre-trained networks for $\hat{y}$ and $y$ \cite{yu2020multimodal}, to provide additional supervision to the generator. In face-swapping methods the generator of the GAN framework receives the frames of the target video and a representation (i.e image or video) of the source identity. Then the generator learns to map the attributes of the source face onto the target face. Similarly, in most literature that leverages the GAN framework for face reenactment, the source image and the driving modality (i.e. audio, video, and text) are encoded and fused in the encoder of the generator. The decoder learns to map this encoded representation into the target frames. 

\textit{Cycle GAN:} The success of the Cycle GAN model proposed by Zhu et al. \cite{zhu2017unpaired} in unpaired Image-to-Image Translation has also been leveraged in several face reenactment literature \cite{wu2018reenactgan, xu2017face}. Specifically, the Cycle GAN removes the need for paired inputs and targets from input, $X$, and target, $Y$ domains using a cycle consistency, which ensures that translating from one domain to another and then back results in an output close to the original. Formally, the cycle consistency loss can be written as,
\begin{equation}
\begin{split}
  L_{Cycle-GAN} & = \min_{G_X, G_Y} \max_{D_X, D_Y} \mathcal{L}_{\text{GAN}}(G_X, D_X, Y, X) \\ & + \mathcal{L}_{\text{GAN}}(G_Y, D_Y, X, Y) + \lambda \mathcal{L}_{\text{cycle}}(G_X, G_Y),
\end{split}
\end{equation}

where $G_X$ and $G_Y$ denote generators for domains X and Y, while 
$D_X$ and $D_Y$ are the discriminators for domains X and Y, respectively. $\lambda$ is a hyperparameter controlling the importance of cycle-consistency. The works that utilises the Cycle GAN framework for deepfake face generation follow a similar approach to GAN-based architectures. These works treat this encoded representation as domain X and the target video as domain Y. The main difference between GAN-based approaches and Cycle GAN-based approaches is the requirement for paired inputs and targets requirement in GAN-based approaches while Cycle GAN allows many input source images to be mapped into one target video.

\textit{Recurrent GANs:} Motivated by the need to capture temporal information in the generation process, authors of several works have based their frameworks on recurrent GAN network structure.

Within the structure of the generator and discriminator, the Recurrent GAN utilises recurrent neural networks which allow it to model the temporal relationships within the data. Recurrent GAN possesses two additional losses, in addition to the typical GAN loss, namely, temporal loss and recurrent loss. 

The temporal loss can be written as, 
\begin{equation}
\mathcal{L}_{\text{temporal}}(\textbf{G}) = \frac{1}{T-1}\sum_{t=1}^{T-1} \| \textbf{G}(\textbf{z})_t - \textbf{G}(\textbf{z})_{t+1} \|_2^2
\end{equation}
is designed to promote smoothness in the penalise generated sequence by minimising the difference between consecutive frames. The coherence of the generated sequence is encouraged by the recurrent loss where,
\begin{equation}
\mathcal{L}_{\text{recurrent}}(\textbf{G}) = \frac{1}{T}\sum_{t=1}^{T} \| \textbf{G}(\textbf{z})_t - \textbf{G}(\textbf{z})_{t-1} \|_2^2
\end{equation}
natural temporal progression in a sequence is encouraged.


\textit{Multimodal GANs:} Especially, in audio and text-driven face reenactment scenarios multimodal fusion strategies are employed due to the distinct modality representation of the input space capture. Some of the popular modalities that are used in literature are, audio features \cite{yu2020multimodal,agarwal2023audio}, linguistic features \cite{yu2020multimodal, fan2022faceformer}, categorical representation of the emotion \cite{goyal2023emotionally}, and expression, pose, and gaze-related features \cite{agarwal2023audio}. 

Numerous fusion strategies ranging from direct concatenation \cite{goyal2023emotionally} to attention-based fusion \cite{yu2020multimodal} have been leveraged in the generator to combine the multimodal inputs.

\textbf{Latent Space Decomposition:} A latent space is a learned lower-dimensional representation of the data that captures only the essential features. Several architectures, including, Variational Autoencoders, and GANs utilise the concept of latent space transfer which involves manipulating this learned latent space such that the decoded representation is of another representation of the input. For instance, \cite{wu2018reenactgan} proposes to map the input visual representations into a boundary latent space that represents a face with respect to its boundaries instead of raw pixels-based values. Several works \cite{wang2021one,jang2023s} within the face deepfakes generation literature have extended the concept of latent space transfer such that the encoded features in the latent space are decomposed into sub-attributes or sub-regions in which the features are disentangled. 
For example, in \cite{jang2023s} the authors propose a decomposition of the audio and visual inputs into canonical space and multimodal motion Space. In canonical space, every face has the same motion patterns but different identities while in the multimodal motion space only represents motion-related features irrespective of identities. 

\textbf{Diffusion Models:} Diffusion models also operate based on the concept of manipulating the latent space. They extend this concept by learning the underlying probability distribution of the data in the lower-dimensional space. Furthermore, they often employ hierarchical structures to capture the latent space in multiple levels of abstraction in the data, facilitating learning of both local and global structures. Diffusion models are newly emerging within the landscape of deepfake generation \cite{chen2023text, xu2023multimodal, wang2022diffusion}. They are preferred over the GAN-based counterparts due to the stability of training of diffusion models over GANs. Generators of the GAN often suffer from issues such as mode collapse \cite{xu2023multimodal}. On the other hand diffusion models exhibit more stable training dynamics due to their well-defined learning objectives.

Among different variants of diffusion models denoising diffusion models are most commonly used. Specifically, a diffusion process gradually adds noise to the data sampled from the target distribution as a Markov chain and the objective is to estimate the clean version of the input by leveraging the reverse diffusion process. This objective could be written as,
\begin{equation}
 \mathcal{L}_{\text{denoise}} = \mathbb{E}_{\mathbf{x} \sim p_{\text{data}}} \left[ \frac{1}{2\sigma^2} \| \mathbf{x} - \tilde{\mathbf{x}} \|_2^2 + \frac{1}{2} \log(2\pi\sigma^2) \right],   
\end{equation}
where $\mathbf{x}$ is the ground truth clean data sample, $\tilde{\mathbf{x}}$ is the denoised data sample generated by the diffusion model. $\sigma^2$ denotes the variance of the noise added to input during the forward diffusion process and $\|.,.\|_2^2$ is the Euclidean distance between two vectors $\mathbf{x}$ and $\tilde{\mathbf{x}}$. Once the training of the diffusion model completes the learned latent space is used for facial manipulation.

\subsubsection{Performance evaluation} 

Several different metrics have been used to measure the quality of the generated video. Among these methods L1, LMD, AED, ID, PSNR, SSIM, EFD and Sync are popular. \textbf{L1} measures the average L1 distance between the ground truth and generated video considering all the pixels while \textbf{LMD} measures the distance with respect to facial landmarks using a pre-trained facial landmark detector \cite{siarohin2019first}. \textbf{AED}  Average Euclidean Distance measures the distance of the ground truth and generated behavioural information using a pre-trained facial feature extractor such as Openface \cite{amos2016openface}. Among the extracted behavioural features expression, face angle, and eye gaze are popular. Several works have also used distance in terms of identity \textbf{(ID)} features extracted from pre-trained face recognition models Curricularface \cite{huang2020curricularface} and Arcface \cite{deng2019arcface} to measure the quality of the synthesised faces. Typically this distance is measured as cosine similarity between the ground truth and generated identity features. Peak Signal to Noise Ratio \textbf{(PSNR)} evaluates the reconstruction quality of the generated image sequence compared to the ground truth image sequence. Similarly, the Structural Similarity Index 
 \textbf{(SSIM)} has been used to evaluate the changes with reference to the structural information of the ground truth and generated images. To measure the ability of the deepfake generation methods to synthesise natural human emotions some works \cite{xu2023multimodal, goyal2023emotionally} employ pre-trained emotion recognition models such as Affectnet \cite{mollahosseini2017affectnet} and measure Emotion/Expression Feature Distance (EFD). To measure the lip synchronisation \textbf{(Sync)} several works \cite{fan2022faceformer, xu2023multimodal, jang2023s} have used the Syncnet \cite{chung2017out} or Meshtalk \cite{richard2021meshtalk} confidence score. In addition, we would like to point out that a few studies have used human trials to validate the perceptual realism of the generated synthetic media. For instance, in \cite{wu2018reenactgan} 30 volunteers are used to compare the quality of the images generated by \cite{wu2018reenactgan} and baseline methods using the protocol presented in \cite{isola2017image}

\subsubsection{Literature review on deepfake generation}
\begin{figure}
    \centering
    \includegraphics[width=.8\linewidth]{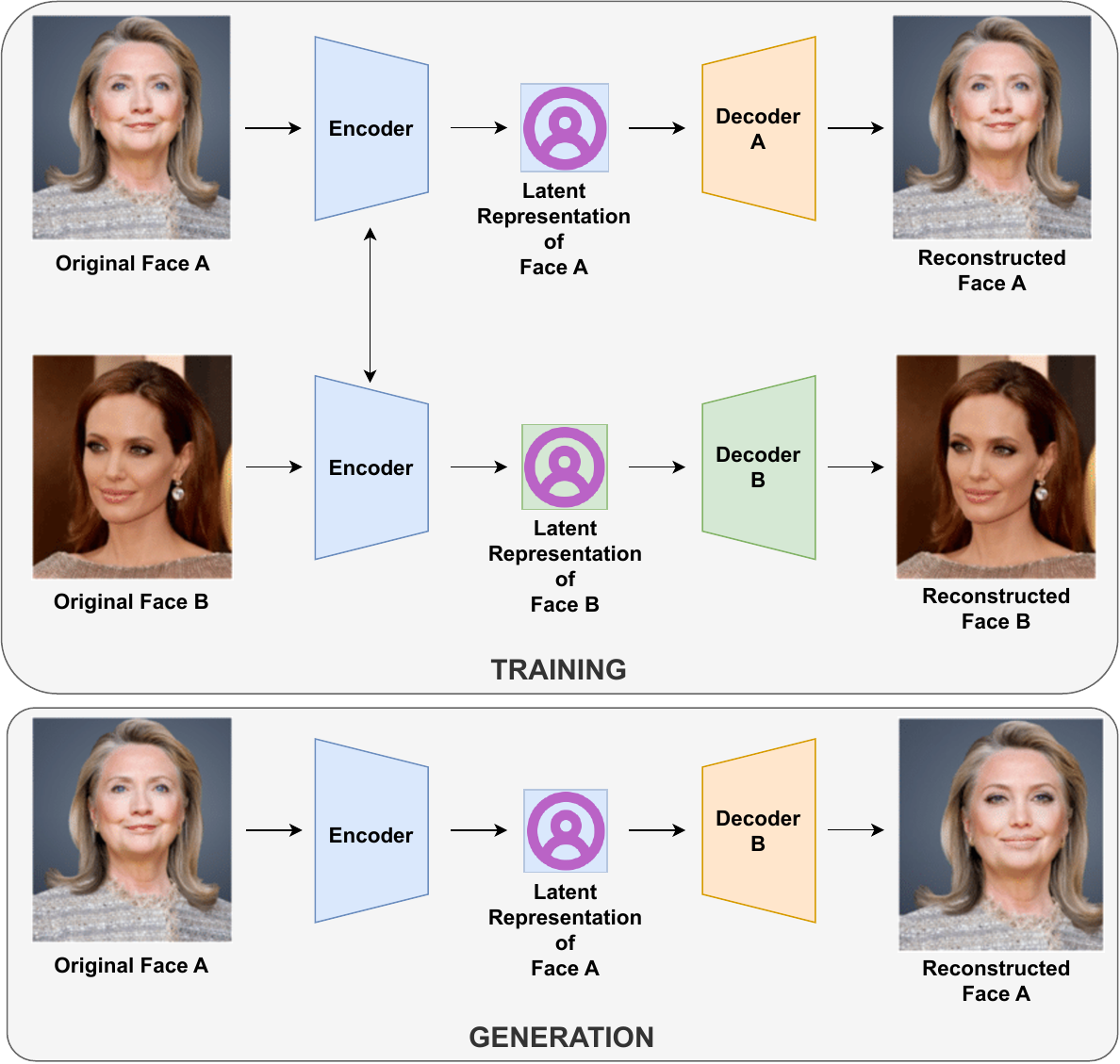}
    \caption{Illustration of the autoencoder-based framework introduced for face swapping. A shared encoder generates latent representations for source and target faces and the two decoder networks recreate the inputs of their respective identities. In the face-swapping stage, the decoder of the source face is used to reconstruct the source face on the target video.}
    \label{fig:ae_face_swapping}
\end{figure}

This section presents the literature review of deepfake generation technology under two categories, face swapping and face reenactment. \newline \newline
\textbf{Face Swapping:} One of the first approaches for successfully generating face swap videos is from 2017 when a Reddit user utilised an autoencoder-based framework to generate face deepfakes \cite{waseem2023deepfake}. During the training phase of this architecture, which is illustrated in Fig. \ref{fig:ae_face_swapping}, it receives images of two separate identities and a shared encoder generates latent representations for those two inputs. The motivation for using a shared encoder is to learn a shared latent space to represent both source and target individuals. Two decoder networks are used to recreate the inputs of their respective identities. Once the training completes the decoder of the source face is leveraged to reconstruct the source face on the input frames of the target video. Due to the simplicity and powerfulness of this framework numerous applications, including, DeepFaceLab and DFaker, have been proposed which are based on the principles of this autoencoder architecture.

A landmark detection-based approach is proposed in \cite{nirkin2018face} where the authors propose to first detect 2D facial landmarks in both source and target faces to compute the 3D pose which accounts for both viewpoint and expression of the respective faces. Then the face regions are segmented using a pre-trained fully convolution neural network (FCN) to remove background and occlusions. During the face transfer stage, the source face is warped onto the target face using the alignment priors computed based on the 3D face poses. As the final step, the authors propose to blend the overlayed source face with the target face's background using an off-the-shelf algorithm \cite{perez2003poisson}. Despite the significant advances made by these methods, their perceptual quality was poor as they left a lot of artifacts in the generated faces. To account for those limitations and to improve the quality of face swapping GAN-based approaches were proposed. 

Early GAN-based methods such as Face-swap GAN (FS-GAN) \cite{deepswapgan} and DeepFaceLab \cite{perov2020deepfacelab} require subject-specific training. For instance, in the FS-GAN method, the encoder-decoder networks of the autoencoder architecture are used as the generator of the GAN framework, and a discriminator is added to provide additional supervision. In addition to this adversarial loss, the mean square error between the reconstructed and the ground truth face and perceptual loss which is computed using the features extracted using the VGGFace model are used to guide the generator training. Due to the subject specificity of the training, the capabilities models are restricted for swapping faces between specific identities \cite{waseem2023deepfake}, hence offering limited generalistion ability. 

\begin{figure*}[htbp]
    \centering
    \includegraphics[width=\textwidth]{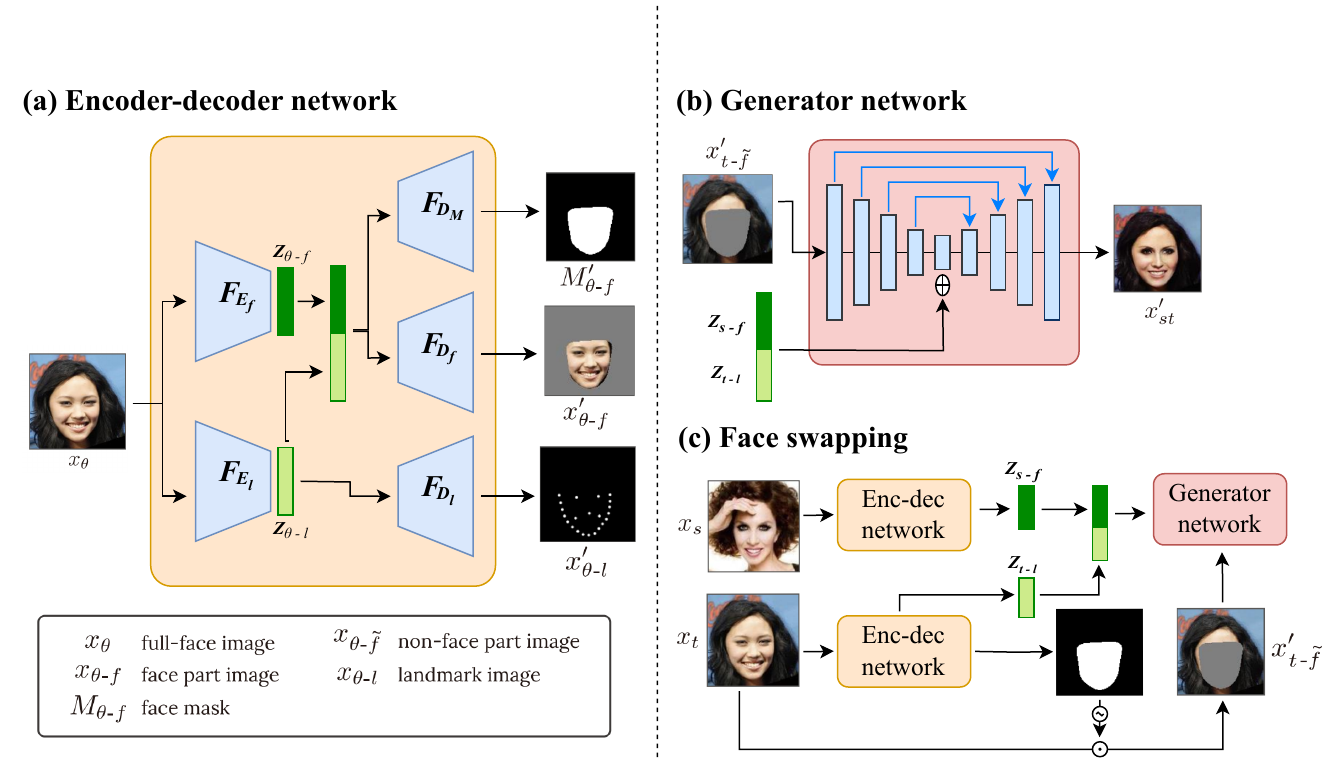}
    \caption{Illustration of the architecture of FSNet model \cite{natsume2019fsnet} which is composed of VAE-based encoder-decoder architecture, and a GAN based generator network.}
    \label{fig:FSNet}
\end{figure*}

Subject-agnostic methods have emerged to overcome the limitations of subject-specific approaches. FSNet \cite{natsume2019fsnet} and RSGAN \cite{natsume2018rsgan} are two popular subject-agnostic methods that are based on VAE. Specifically, FSNet employs a VAE-based encoder-decoder architecture and obtains a latent representation of the face that is independent of the face geometry and appearance of the non-face region in the image. Then the generator of the GAN framework leverages this latent variable and synthesizes a face-swapped image. Fig. \ref{fig:FSNet} visually illustrates this framework. When training this framework the authors have used two separate losses for training the VAE and the GAN. The VAE generates three outputs, namely, face mask, face-part image, and landmark image. The authors propose to use cross entropy losses for the face masks and landmark images and an L1 loss for the face-part images. In addition, identity loss which is calculated as triplet loss is also utilised in training. The proposed GAN framework has two discriminator networks, a global discriminator that distinguishes real and synthesised images, and a patch discriminator that classifies whether a local patch of the image is from a real or a synthesized image. These two discriminators generate two adversarial losses to govern the generator network. In contrast, the RSGAN framework comprises three sub-networks, two separator networks, and a composer network. The separator networks generate latent space representations for face and hair regions of the input image and the composer is trained to reconstruct the input face image using these latent space representations. Similar to FSNet global and patch discriminator networks are used in this architecture. For training the VAE, three reconstruction losses are defined, representing the reconstruction of the face region, hair region, and the full image. For GAN training, similar to FSNet, two adversarial losses based on the two discriminators are incorporated. To enforce the composer network to learn the visual attributes, the authors have added a classifier to the composer which classifies the visual attributes of the input.

Version 2 of the Face Swapping GAN (FS-GANv2) \cite{nirkin2022fsganv2} is a GAN-based architecture that is capable of face swapping and face reenactment. The iterative architecture of FSGAN enables it to handle occluded face regions using interpolation. This architecture is composed of three main components, a reenactment generator and the segmentation CNN module, a face inpainting network, and a blending module. The reenactment generator and the segmentation CNN module receives facial landmarks of the target face such that the pose and expression of the source face can be augmented to match the target face. Then the segmentation CNN computes the segmentation masks for the hair and face of this augmented source face. The face inpainting network inpaints the missing parts due to occlusions and the blending network blends the swapped face region to match the illumination and skin tone of the target face. When training this GAN framework perceptual loss computed based on VGG-19 \cite{simonyan2014very}, reconstruction loss computed using L1 loss, and adversarial loss computed using the multi-scale discriminator architecture of pix2pixHD \cite{wang2018high} is used.

In a different line of work, the authors of the FaceShifter \cite{li2019faceshifter} model propose to extensively utilise target face information in the face-swapping process. Specifically, an attributes encoder for extracting multi-level target face attributes in various spatial resolutions is leveraged. Therefore, the identity encoder of this framework encodes the identity information of the source image in latent space while the attributes encoder receives the target face and extracts attributes of the target. Leveraging these two latent embeddings the proposed Adaptive Attentional Denormalization (AAD) Generator generates the swapped face image. For training this framework the authors have utilised a series of losses, including, adversarial loss computed using the multi-scale discriminator architecture,  identity preservation loss computed using cosine similarity loss, attributes preservation loss at the embedding level, and reconstruction loss as pixel level computed as L2 distance. 

Another notable GAN-based architecture in face swapping is the SimSwap \cite{chen2020simswap} model which proposes an architecture that is generalisable to arbitrary faces and preserves the facial expression and gaze direction of the target face during the swapping process. One of the pivotal contributions of this work is the introduction of the ID injection module which transfers the identity information of the source face into a feature representation that the decoder uses in the decoding process. In addition, an identity loss that encourages this translation to have a similar identity as the source face and a weak feature matching Loss that preserves the attributes of the target face are proposed to train the network. Therefore in total the training process of SimSwap leverages, adversarial loss, reconstruction loss, identity loss, and weak feature matching loss. 

The latent space disentanglement approach of StyleGAN has inspired a few face-swapping architectures. For instance, Xu et al. \cite{xu2022high} proposed an approach that disentangles the texture and appearance features of the source and target faces. During the face-swapping process, the source identity and texture characteristics are mapped to the target appearance features. To maintain the face structure facial landmarks of the source and target faces are also encoded. The authors have utilised a series of loss functions for training this architecture which includes, adversarial loss,  identity-preservation loss, landmark-alignment loss, and style-transfer loss which is based on BeautyGAN \cite{li2018beautygan}. The authors of MegaFS \cite{zhu2021one} utilises the latent space of StyleGAN2 for high-resolution face swapping with different identities. The proposed
Hierarchical Representation Face Encoder (HieRFE) \cite{xu2022high} encodes the facial attributes in a hierarchical manner to maintain more facial details. This is achieved via a ResNet50 backbone-based multiple residual blocks for the extraction of salient features and representing these in a feature pyramid structure based on Feature Pyramid Network \cite{lin2017feature}. In the next stage, the Face Transfer Module (FTM) which controls the mixing of the latent spaces of the source and target faces. For training the HieRFE pixel-wise reconstruction loss, perceptual loss, identity loss, and landmarks loss are used. For FTM the authors propose to use the above four losses as well as a stabilisation loss to stabilise the training process. 

More recently, \cite{rosberg2023facedancer} proposes the FaceDancer architecture to overcome the challenges that the existing methods face due to the lighting, occlusions, and pose variations in the source and target faces. This paper introduces two major modules: Adaptive Feature Fusion Attention (AFFA) and Interpreted Feature Similarity Regularization (IFSR). The AFFA produces attention masks to gate the incoming features that have been conditioned on the source identity information and the unconditioned target face information selectively. Specifically, during the training process AFFA learns which conditioned features (e.g. identity information of the source face) to discard and which unconditioned features (e.g. background information) to keep in the target face. In contrast, the IFSR is proposed for preservation of the attributes such as facial expression, head pose, and lighting while still transferring the identity. The authors employ a series of loss functions during the training, including, identity loss, reconstruction loss, perceptual loss, cycle consistence loss, and adversarial loss.

\noindent \textbf{Face Reenactment:} One of the pioneer works in the domain of face reenactment is the Face2Face project \cite{thies2016face2face}.  The facial expressions of both source and target video are tracked. The mouth interior that best matches the re-targeted expression is retrieved from the target sequence and warped to produce an accurate fit. Finally, a blending process is conducted to seamlessly blend the new expression on the target face. However, it should be noted that the coarse 3D facial reconstructions of the target face make the reconstructed target face not accurately follow the source person’s head and eye movements \cite{waseem2023deepfake}. When training this framework the authors have utilised photo-metric alignment loss at a pixel level which measures how well the synthesised image represents input data, a feature alignment loss that enforces feature similarity between a set of salient facial feature points, regularisation loss to make the synthesised faces follow a normal distribution are used.

In a different line of work, ReenactGAN \cite{wu2018reenactgan} utilises the concept of facial boundary transfer for the task of face reenactment. The authors show that the direct transfer of facial movements and expressions at the pixel level is suboptimal and could result in structural artifacts. As such, the authors propose to map the source face into a boundary latent space, and a transformer is subsequently used to adapt the source face’s boundary to the target’s boundary in this latent space. The authors show that learning in boundary space allows them to perform model training without paired data, enabling them to perform many-to-one mapping such that they can reenact a target face based on images or videos from arbitrary sources. For training this framework the authors have used cycle consistency loss, adversarial loss, and a shape constraint loss that encourages a transformed boundary to better follow its source.

An approach that utilises emotion Action Units (AU) for generating diverse facial expressions is proposed in GANimation \cite{pumarola2018ganimation}. The authors show that popular GAN-based approaches suffer from the inability to generate diverse expressions and are limited to generating a discrete number of expressions, determined by the content of the dataset. In contrast, the AU-based approach allows conditioning the GAN synthesis using a continuous manifold allowing them to control the magnitude of activation of each AU. For training this pipeline adversarial loss, total variation loss that is computed as the sum of the squared differences for neighboring pixel values, conditional expression loss that guides the generator to generate target expression, and identity loss are used. 

One-shot and few-shot learning methods have emerged to overcome the need for large-scale datasets of source and target identities for training the existing models which makes them ineffective for reenacting unknown identities. Among the few-shot learning models, \cite{siarohin2019first} and \cite{zakharov2019few} are notable considering the robustness they achieve in diverse settings. Specifically, the First Order Motion Model (FOMM) \cite{siarohin2019first} proposes a few-short learning architecture that decouples appearance and motion information using a self-supervised learning objective.  To account for complex motions, the motion components around the learned keypoints are represented with their local affine transformations. This few-slot learning framework is capable of generating re-enactments with just a few training examples. Furthermore, the generator of this network is capable of handling occlusions using an occlusion mask for regions not visible in the source image and anticipating their appearance. To train this framework the authors have used a reconstruction loss based on the perceptual loss, a relative motion transfer loss, and a keypoint localisation loss.

Zakharov et al. \cite{zakharov2019few} proposed a few-shot learning architecture that is based on meta-learning. This architecture is composed of three main components, namely the embedder network maps the input images to the embedding vectors, a generator network maps input face landmarks into output frames, and a discriminator to determine the realism of the syntheised images. During the meta-learning stage, the authors trained all three subcomponents of their framework using content loss, adversarial loss, and embedding match loss. The content loss measures the distance between the ground truth image and the reconstruction using the perceptual similarity measure. The embedding match loss encourages the similarity of the two types of ground truth and the encoded image. 

\begin{figure}[htbp]
    \centering
    \includegraphics[width=\linewidth]{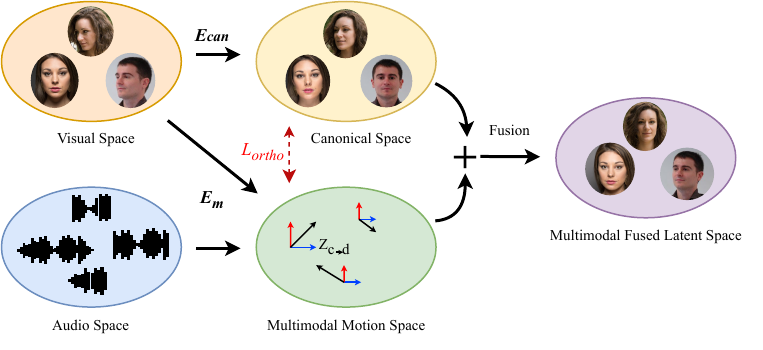}
    \caption{Disentanglement of the canonical and multimodal motion latent spaces in \cite{jang2023s} which allowed them to manipulate only the motion-related features and preserve identity features.}
    \label{fig:jang2023s}
\end{figure}

Audio-driven facial reenactment has recently attained significant traction within the research community due to its numerous applications ranging from virtual assistants to dubbing. One of the pioneering works within this domain is in \cite{jang2023s} where the authors propose a framework that is highly flexible in terms of accounting for full target motion including head pose, eyebrows, eye blinks, and eye gaze movements. The authors achieve it by manipulating the motion-related latent space of the face while preserving semantically meaningful features associated with the identity. Specifically, this framework disentangles the latent space of StyleGAN into two distinct subspaces, (i) canonical space that captures different facial identities irrespective of facial attributes, and (ii) multimodal motion space that contains motion features irrespective of modality. The disentanglement of the two subspaces is achieved via introducing an orthogonality constraint between the canonical space and the multimodal motion space. To train this framework the authors have utilised a series of loss functions, including, adversarial loss, identity loss, L1 reconstruction loss, perceptual loss, synchronisation loss which is formulated using SyncNet \cite{chung2017out}, and orthogonality loss implemented by extending \cite{yang2021l2m}. This architecture is visually depicted in Fig. \ref{fig:jang2023s}. 

The authors of \cite{yu2020multimodal} illustrated the importance of learning the correlation between speech and the movement of the face region around the mouth (lips, cheeks, and chin) and proposed a novel speech-driven face reenactment architecture named Face2Vid. In this architecture, a time-delayed LSTM that receives both text and audio inputs is adopted to predict mouth landmarks. Leveraging these landmarks in the next module generates optical flow frames such that smooth transitions in both lips and facial movements can be achieved throughout the entire synthesised video clip. Finally, the Face2Vid module translates these optical flow images into video frames. To train this framework the authors have employed, a temporal adversarial loss, a feature mapping loss based on the discriminators \cite{wang2018high}, and the perceptual loss. 

In a different line of work, the authors of \cite{goyal2023emotionally} argue that little focus has been paid to people's expressions and emotions when synthesising deepfake faces and proposes a conditional GAN framework that is capable of generating more realistic and convincing videos with a broad range of six emotions, including, happiness, sadness, fear, anger, disgust, and neutral. This architecture extends the Wav2Lip \cite{chung2017out} framework by conditioning the synthesis on the categorical one-hot vector representation of the emotion and by using an additional emotion encoder and an emotion discriminator. Similar to the prior work the authors have utilised L1 reconstruction loss, and perceptual loss for training this framework. In addition, lip sync-loss loss and emotion discriminator loss are used to provide further guidance.

More recently, Agarwal et al. \cite{agarwal2023audio} have proposed an Audio-Visual Face Re-enactment GAN named AVFR-GAN. In contrast to purely video or audio-driven architecture, the AVFR-GAN uses both audio and visual cues to generate highly realistic face reenactments. When encoding face information the authors propose to provide additional priors about the structure of the face in the form of a face segmentation mask and face mesh. Melspectrogram representation of the speech is also provided to an audio encoder to help reenact the mouth region. The extracted audio and visual feature maps are combined, warped, and passed to the identity-aware generator along with the source face which generates the reenacted frame. L1 reconstruction loss, perceptual similarity loss, and the equivarience constraints of \cite{siarohin2019first} are the loss functions used for training AVFR-GAN.

\noindent\textbf{Summary of face deepfake generation methods and open research questions:} GAN-based technologies have been the dominant approach for the generation of deepfakes within both face swap and face reenactment categories. Preservation of facial expressions, head-poses, etc. of the target face has been one of the main challenges within the face swapping task. MegaFS \cite{zhu2021one}, FaceDancer \cite{rosberg2023facedancer} architectures take an important step towards the preservation of target facial attributes, however, still there exist limitations with respect to handling occlusions and the resolution of the synthesised faces. On the other hand, the recent trends within the face reenactment literature have been on audio-driven facial reenactment and few-slot learning frameworks that could be trained with a few training examples. While these attempts are commendable and could elevate the utility of face deepfake generation technology with numerous novel applications, the synthesised faces lack realism. For instance, most of the recent state-of-the-art methods such as FC-TFG \cite{jang2023s}, AVFR-GAN \cite{agarwal2023audio}, and GANimation \cite{pumarola2018ganimation} within the face reenactment category are only capable of generating face deepfakes at $\leqq 256 \times 256$ resolution. This is significantly lower when compared with the output resolution of the recent face swap technology such as MegaFS \cite{zhu2021one} and HiRFS \cite{xu2022high} which can achieve resolution up to $1024 \times 1024$. Furthermore, most of the existing state-of-the-art methods within the face reenactment domain lack gaze adaptation, cannot handle extreme poses, and fail to preserve source facial features. In Tabs. 1 and 2 in supplementary material we summarise the state-of-the-art face-swap and face-reenact deepfake generation methodologies, respectively, and discuss their strengths and weaknesses. 

\noindent\textbf{Top-ten tools for generating face deepfakes:}
In Tab. \ref{tab:face_get_tools} we summarise top-ten tools, including free and open source tools, that are available for the generation of face deepfakes. When comparing the available tools for ranking we consider the quality of the generated faces, customizability, types of media that they can manipulate and the availability of the source codes. 
\begin{table*}[htbp]
\caption{Top ten tools to create face deepfakes.}
\resizebox{\linewidth}{!}{%
\begin{tabular}{|c|c|c|c|c|c|c|}
\hline
deepfake Type                    & Method                   & Images & Audio & Video & Open-source & URL                                                       \\ \hline
\multirow{6}{*}{Face Swap}        & Face Swap Live  & \xmark      & \xmark     & \cmark     & \xmark           & https://apps.apple.com/us/app/face-swap-live/id1042987645 \\ \cline{2-7} 
                                  & Deepfakes Web            & \xmark      & \xmark     & \cmark     & \xmark           & https://deepfakesweb.com/                                 \\ \cline{2-7} 
                                  & FaceMagic                & \cmark      & \xmark     & \cmark     & \xmark           & https://www.facemagic.net/faceswap                        \\ \cline{2-7} 
                                  & DeepFaceLab              & \cmark      & \xmark     & \cmark     & \cmark           & https://github.com/iperov/DeepFaceLab                     \\ \cline{2-7} 
                                  & ReFace                   & \cmark      & \xmark     & \cmark     & \xmark           & https://www.reflect.tech/                                 \\ \cline{2-7} 
                                  & Faceswap                 & \cmark      & \xmark     & \cmark     & \cmark           & https://github.com/deepfakes/faceswap                     \\ \hline
\multirow{4}{*}{Face Reenactment} & Avatarify                & \cmark      & \xmark     & \cmark     & \xmark           & https://avatarify.ai/                                     \\ \cline{2-7} 
                                  & Wav2Lip                  & \xmark      & \xmark     & \cmark     & \cmark           & https://github.com/Rudrabha/Wav2Lip                       \\ \cline{2-7} 
                                  & Myheritage               & \xmark      & \xmark     & \cmark     & \xmark           & https://www.myheritage.com/deep-nostalgia                 \\ \cline{2-7} 
                                  & First Order Motion Model & \cmark      & \xmark     & \cmark     & \cmark           & https://github.com/AliaksandrSiarohin/first-order-model   \\ \hline
\end{tabular}}
\label{tab:face_get_tools}
\end{table*}

We would also like to note popular generative AI tools such as Deepbrain
\footnote{https://www.deepbrain.io/}, Midjourney \footnote{http://www.midjourney.com/home} and DALL-E 2 \footnote{https://openai.com/index/dall-e-2/} which are models that are capable of generating realistic images and video from text descriptions. However, these methodologies do not directly fall within the scope of this paper as they are not deepfake tools. Therefore, we do not include them in the comparison in Tab. \ref{tab:face_get_tools}.

\hspace{2mm}
\subsection{Detection}
\subsubsection{Features used for deepfake detection}
\noindent\textbf{Hand-crafted features based approaches:} Early works of deepfake detection leveraged statistical features that have been hand-crafted by analysing the image's pixel values. For instance, \cite{koopman2018detection} proposed the use of Photo Response Non-Uniformity (PRNU) analysis to detect unique noise patterns that are left in the image due to manufacturing defects in the camera sensor. The authors show that when performing the face swap it alters the PRNU patterns of the original video. In another work \cite{kharbat2019image} edge features such as Histogram of Gradient (HoG) features were extracted to illustrate that the edge features from a real video are more correlated than the edge features from the fake video. Xia et al. \cite{xia2022towards} proposed to do a statistical analysis of the colour space of the image to determine the differences between real and fake video frames in various color channels. The input RGB frames are converted to HSV and YCbCr color spaces and first-order differential operators are employed to extract the texture difference features from the colour channels. Wang et al. \cite{wang2022ffr_fd} argued that due to the smoothing of the face region during the blending stage of the face-swapping process, fake videos have fewer feature points than real videos. The authors proposed to use feature point descriptors such as Speeded Up Robust Features (SURF), Scale-Invariant Feature Transform (SIFT), and Oriented Fast and Rotated Brief (ORB) from eight different facial regions such as mouth, inner mouth, eyebrows, eye, and nose. 

\noindent\textbf{Artefacts based approaches:} Majority of the works within the face deepfake detection domain are developed based on detecting the artifacts that are left by the face deepfake generation methods. These artifacts could be broadly categorised into (i) visual artifacts such as facial artifacts, texture artifacts, and boundary artifacts; and (ii) biological artifacts such as changes in heart rate, eye movements, facial movements, and facial expressions.

\noindent\textit{Visual artifacts:} Yang et al. \cite{yang2019exposing} proposed to detect deepfakes via detecting inconsistencies in 3D head pose. In another work \cite{li2021exploiting} facial symmetry is used as the feature for detecting deepfakes. The authors show the existence of inconsistencies and unnatural traces in facial symmetry in face deepfakes. Xu \cite{xu2021deepfake} proposed to use Gray-Level Co-occurrence Matrix to extract texture features with the hypothesis that deepfake generation models produce blurred and irregular textures. In another study \cite{kingra2022lbpnet}, the Local Binary Pattern (LBP) texture feature is used as a descriptor for authenticity. Several methods have also emerged that utilises features from the frequency domain. For instance, in \cite{durall2019unmasking, liu2021spatial} Discrete Fourier Transform (DFT) is used to examine the spectral distributions in both real and fake videos, and in \cite{kohli2021detecting} the inputs are converted into the frequency domain using a 2D global Discrete Cosine Transform and analysed using a CNN. Deep learning models have also been used in some works for extracting artifacts. For instance, Yuezun et al. \cite{li2018exposing} used pre-trained  VGG16, ResNet50, ResNet101, and ResNet152 feature extractors and Kim et al. \cite{kim2021exposing} proposed the use of two deep feature extractors to simultaneously extract content features and trace features from a face image. In \cite{agarwal2020detecting} the authors propose the use of appearance and motion features of the face extracted from pre-trained VGG-19 and Facial Attributes-Net \cite{wiles2018self} models, respectively.

\noindent\textit{Biological artifacts:} Agarwal et al. \cite{agarwal2019protecting} argued that the face deepfakes lack the expressiveness of real videos and proposed a method that is based on analysing the movement of facial landmarks. This approach is extended in \cite{agarwal2020detecting} to capture head poses, facial landmarks, and expressions. In a different line of work \cite{nguyen2020eyebrow} inconsistencies in the eyebrow region are extracted and analysed using CNN-based feature extractors including ResNet and SqueezeNet \cite{iandola2016squeezenet}. Lip reading models have been leveraged by Haliassos et al. \cite{haliassos2021lips} to map irregularities in mouth movement and \cite{liao2023famm} proposes to examine facial muscle motion features. In addition, the lack of eye blinks has also been leveraged as a biological artifact in face deepfakes. Heart rate estimation-based features have also been utilised in the literature. For instance, irregularities in heartbeat rhythms from blood flow in the face \cite{qi2020deeprhythm} and inconsistencies in colour changes in the face caused by variations in oxygen concentration in the blood \cite{fernandes2019predicting, hernandez2020deepfakeson} have been used. Several remote PhotoPlethysmoGraphy (rPPG) based methods \cite{ciftci2020fakecatcher, wu2024local} have also emerged to detect face deepfakes by analysing the irregularities in light absorption in facial skin tissues. 

\noindent\textbf{Deep Learning based approaches:} Under deep learning-based approaches we categorise the neural network architectures that have been proposed to learn features that are indicative of face deepfakes automatically, instead of focusing on a particular feature as the methods mentioned earlier in this section. MesoNet \cite{afchar2018mesonet} and Capsule Network architecture of \cite{nguyen2019capsule} are some of the early works within this domain. However, these models poorly generalise to unseen deepfake generation methods \cite{waseem2023deepfake}. Motivated by the fact that real images have consistent source characteristics throughout the image while manipulated images have inconsistent source characteristics Zhao et al.  \cite{zhao2021learning} proposed a model that is trained using pair-wise self-consistency learning paradigm. An autoencoder-based approach is proposed in \cite{cozzolino2018forensictransfer} which is capable of generalising different yet related manipulation methods. This is achieved via learning a compact embedding that could be translated between different manipulation domains by activating specific regions of the latent space. Kumar et al. \cite{kumar2020detecting} proposed training separate CNN feature extractors to specific facial regions and proposed a framework consisting of five ResNet-18 models. An ensemble of deep learning networks that incorporates numerous state-of-the-art classification models, including XceptionNet, MobileNet, ResNet101, InceptionV3, DensNet121, InceptionReseNetV2, and DenseNet169, into a single pipeline is proposed in \cite{rana2020deepfakestack}. 

Spatio-temporal deep learning approaches are also popular within the face deepfakes detection literature as they possess the ability to analyse spatial and temporal consistency across video frames. Guera \cite{guera2018deepfake} proposed to analyse the framewise feature of a video extracted from a CNN using an LSTM network. A joint learning framework where the CNN architecture is jointly trained with an RNN is proposed in \cite{sabir2019recurrent}. A 3DCNN is proposed in \cite{nguyen2021learning} to simultaneously process spatial and temporal dimensions. A novel architecture named Interpretable Spatial-Temporal Video Transformer (ISTVT) is proposed in \cite{zhao2023istvt} which leverages  Xception blocks to extract spatial features and the authors propose to map spatial and temporal correlations using self-attention modules. In a similar line of work, vision transformers for face deepfake detection are introduced in \cite{wodajo2021deepfake} where the authors propose a network named  Convolutional Vision-Transformer (CVT). Graph neural networks have also been used within the face deepfakes detection literature where the authors of \cite{shang2023constructing} propose the Spatial Relation Graph Unit (SRGU). This architecture can capture local and global spatial inconsistencies through graph convolution. Features of the same spatial location across different frames are modeled into a fully connected graph and a cosine distance-based similarity matrix to detect temporal incoherencies. 

Contrastive learning-based approaches are also popular within the deep learning-based face deepfakes detection approaches. For instance, Xu et al. \cite{xu2022supervised} supervised contrastive (SupCon) learning to discriminate between real and fake images while \cite{dong2023contrastive} proposes a framework to combine intra-domain and cross-domain formation to improve generalisation. 

When considering the multimodal approaches, Mittal et al. \cite{mittal2020emotions} proposed the utilisation of emotion features extracted from audio and video modalities and analysing the inconsistencies. The architecture of \cite{chugh2020not} is a two-stream architecture where the visual stream uses a 3D- ResNet architecture to extract features, and Mel-Frequency Cepstral Coefficients (MFCC) are extracted as audio features. This framework is trained to detect irregularities within the audio and video modalities in a contrastive manner. The framework of Zhou et al. \cite{zhou2021joint} leverages the concept of temporal alignment between audio and video streams. Two separate subnetworks are used to model video and audio streams separately and a synchronisation stream is used to learn the synchronisation patterns between modalities. 

\noindent\textbf{Anomaly detection based approaches: } The main difference between the deep learning based approaches mentioned above and the anomaly detection based approaches is that deep learning based approaches treat deepfake detection as a classification problem where they classify the input into real or fake classes. In contrast, the anomaly detection based approaches formulate deepfake detection as the task of learning normality and detection of fake media with respect to the deviation from this normality. 

One of the pioneering works within anomaly detection based approaches is in \cite{khodabakhsh2020generalizable} in which the authors propose a probabilistic approach that predicts the logarithmic probability of observing a particular pixel's intensity by considering the relationship between previous pixels. In a different line of work local motion patterns of real videos are analysed in Wang et al. \cite{wang2020exposing} to detect anomalies in fake videos. \cite{khalid2020oc} uses a VAE to reconstruct real images and the fake images are detected by considering the root mean square error between the input and reconstructed image. Audio-visual features of authentic videos learned using the large-scale Voxceleb dataset \cite{nagrani2017voxceleb} are used in \cite{cozzolino2023audio} for detecting anomalies caused by deepfakes.

\subsubsection{Literature review on deepfake detection} 

In this section, we summarise the state-of-the-art face deepfake detection methods under artefact-based approaches, deep learning-based approaches, and anomaly detection-based approaches. Note that we do not include a detailed discussion regarding the hand-crafted feature-based approaches due to their inferior performance in current state-of-the-art benchmarks. For instance, the methods such as \cite{kharbat2019image} and \cite{xia2022towards} methods struggle to detect deepfakes in highly compressed videos, and \cite{wang2022ffr_fd} detecting deepfakes in complex backgrounds \cite{waseem2023deepfake}.

\noindent\textbf{Artefacts based approaches:} The face deepfake detection method of Yang et al. \cite{yang2019exposing} leverages artifacts in 3D head pose. The authors observe that the face-swap algorithms only swap faces in the central face region while keeping the outer contour of the face intact. Due to this mismatch of the landmarks in fake faces, there exist inconsistencies in 3D head pose estimation when it is estimated from central and whole facial landmarks. 68 3D facial landmarks are estimated using the OpenFace2 \cite{baltruvsaitis2016openface} library and the head poses from the central face region and whole face are estimated. The differences between the obtained rotation matrices and translation vectors are used as the features to train a Support Vector Machine (SVM) classifier. In a similar line of work, the movement of facial action units is leveraged in \cite{agarwal2019protecting} for detecting face deepfakes. 16 different facial action units (AU) are extracted using the OpenFace2 library and four additional features, including, pitch and roll of head rotation, the 3D horizontal distance between the corners of the mouth, and the 3D vertical distance between the lower and upper lip are extracted. The authors extract this 20-dimensional feature vector for each frame in a 10-second video and apply Pearson correlation to measure the linearity between these features, yielding a 190-dimensional feature vector which is subsequently fed to an SVM for classification. Nguyen et al. \cite{nguyen2020eyebrow} proposed a biometric matching pipeline for the eyebrow region for the task of detecting deepfakes. Specifically, this framework assumes that a bonafide image of the subject is available and this image is used for biometric enrollment. They evaluated four state-of-the-art deep learning models, including, LightCNN \cite{wu2015lightened}, Resnet, DenseNet \cite{huang2017densely}, and SquezeNet for extracting features for biometric matching.  The cosine distance metric is used to measure the similarity between eyebrow features from the enrolled face and the probe face. 

\begin{figure*}[htbp]
    \centering
    \includegraphics[width=\textwidth]{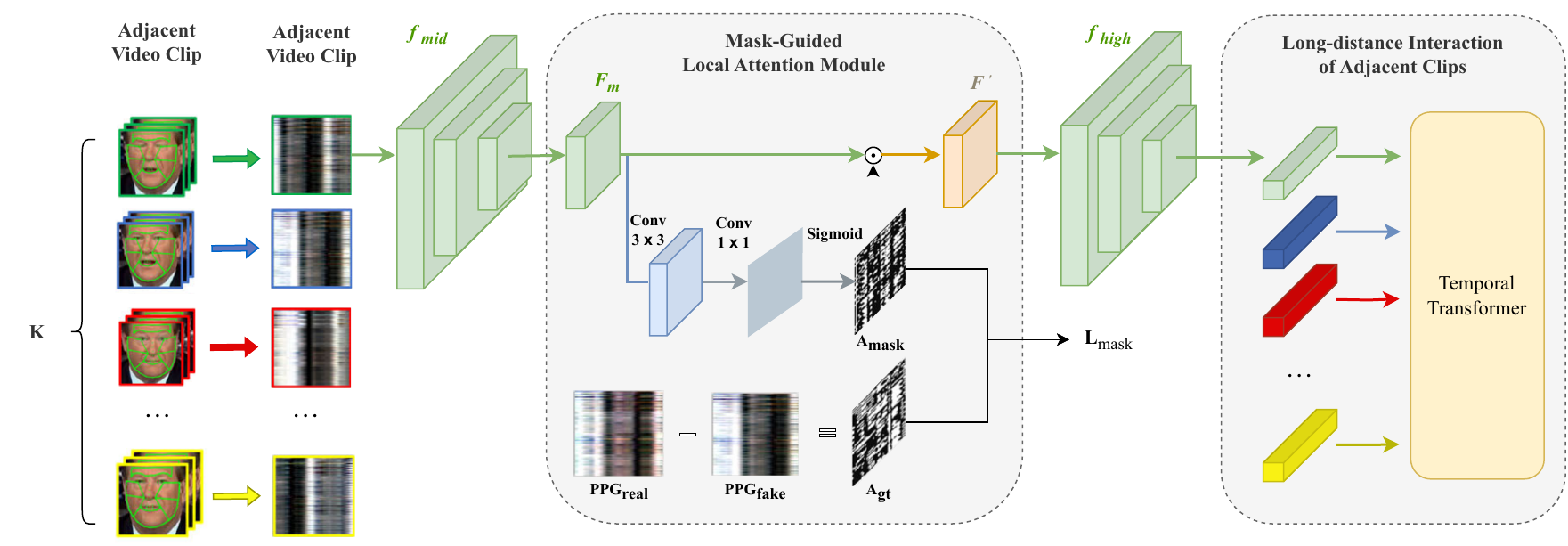}
    \caption{Two-stage network architecture proposed in \cite{wu2024local} which analyses rPPG signals extracted from face, and analyses irregularities in light absorption in facial skin tissues.}
    \label{fig:rppg_model}
\end{figure*}

Physiological measurements such as remote
visual PhotoPlethysmoGraphy (PPG) have also been popular among the assessments for identifying artefacts. Specifically, the DeepRythm architecture of Qi et al. \cite{qi2020deeprhythm} leverages the power of remote PPG which could detect and track the minuscule periodic changes in skin color due to the blood flow through the face from a video. The authors introduce a Motion-Magnified SpatialTemporal Representation (MMSTR) that could capture heart rhythm signals and generate motion-magnified
spatial-temporal map which highlights salient motion regions. The authors also introduce dual spatiotemporal attention to adapt to changing head poses, illumination variations, and different deepfake types. This method has been trained using cross-entropy calculated based on the model's deepfake detection performance. In a similar line of work \cite{ciftci2020fakecatcher} extract G channel-based \cite{zhao2018novel} chrominance-based \cite{de2013robust} remote PPG signals from the left cheek, right cheek, and mid-region of the face. Maps representing the spatiotemporal variations of these signals are constructed which are then used to train a CNN to classify the authenticity. More recently Wu et al. \cite{wu2024local} proposed a two-stage network architecture that could detect the inconsistencies in both spatial and temporal domains of the PPG. This architecture is also illustrated in Fig. \ref{fig:rppg_model}. With the motivation that different video manipulation techniques affect distinct facial regions, the authors first divide the input video into $T$ frame video clips and for each clip face alignment is performed and facial landmarks are obtained. Based on the obtained landmarks sub-regions that encompass cheeks, forehead, and jaw are selected. Average pixel values for each sub-region is computed and the min–max normalisation is applied. These sub-regions are then utilised for the generation of PPG maps. A temporal transformer is employed to capture long-term dependencies between adjacent clips. In addition, a MaskGuided Local Attention module (MLA) is used to highlight the position in the PPG that corresponds to the
modified regions of the face image. To train this network a combination of cross-entropy loss and attention mask loss is leveraged. 

Eyeblink patterns have also been utilised as biological signals for the detection of face deepfakes. For instance, in \cite{li2018exposing} the authors propose a framework to capture the phenomenological and temporal irregularities in eye-blinking that are left by the deepfake generation methods. Specifically, the authors argue that real videos possess periodic eye blinking patterns while the fake videos do not have such blinking patterns. In this pipeline face detection is performed and the face is aligned to a unified coordinate space using facial landmarks. From the aligned face, an area that surrounds the eye is extracted. This region of interest is passed through a CNN to extract features and the temporal relationships across the frames are mapped using an LSTM which predicts the probability of eye blinking. This framework is trained using cross entropy loss. Another work that leverages eye blinking patterns for deepfake detection is in \cite{jung2020deepvision}. This algorithm, named DeepVision takes age, gender, activity and time of the day information in addition to the video for the detection of deepfakes. The eye blink patterns are identified using the Fast-HyperFace \cite{ranjan2017hyperface} for face detection and the Eye Aspect Ratio algorithm \cite{cech2016real} to detect and track the eye. The deepfakes are detected considering the number of eye blinks and the period of blinks. 

\noindent\textbf{Deep Learning Based Approaches: } MesoNet is among the early works that leveraged deep learned features that have been learned end-to-end for the task of detecting face deepfakes. The Meso-4 architecture proposed by the authors is composed of four layers of successive convolutions and pooling followed by two layers of fully connected layers with dropout. Sigmoid activation is used to generate the binary classification. The authors also propose the MesoInception-4 architecture which is generated by replacing the first two convolutional layers of Meso4 by the inception module of \cite{szegedy2015going}. These frameworks were trained using mean squared error loss. The goal of the ForensicTransfer model proposed in \cite{cozzolino2018forensictransfer} is to ensure the generalisation of the model across different but related manipulation types. The authors propose to train an autoencoder-based deep neural network architecture to disentangle real and fake images in the latent space. This training is done on the source domain data. The tuning of the model on the target domain is done using a few target training samples. To train this framework the authors have used both reconstruction loss and the activation loss. The reconstruction loss measures the difference between the input image and the reconstructed image in the pixel space using L1 distance. The activation loss is used to avoid intra-class variations so that there is a clear separation between the latent space of real images and the latent space that corresponds to the images of all manipulation types, including novel manipulation types. 

\begin{figure}[htbp]
    \centering
    \includegraphics[width=\linewidth]{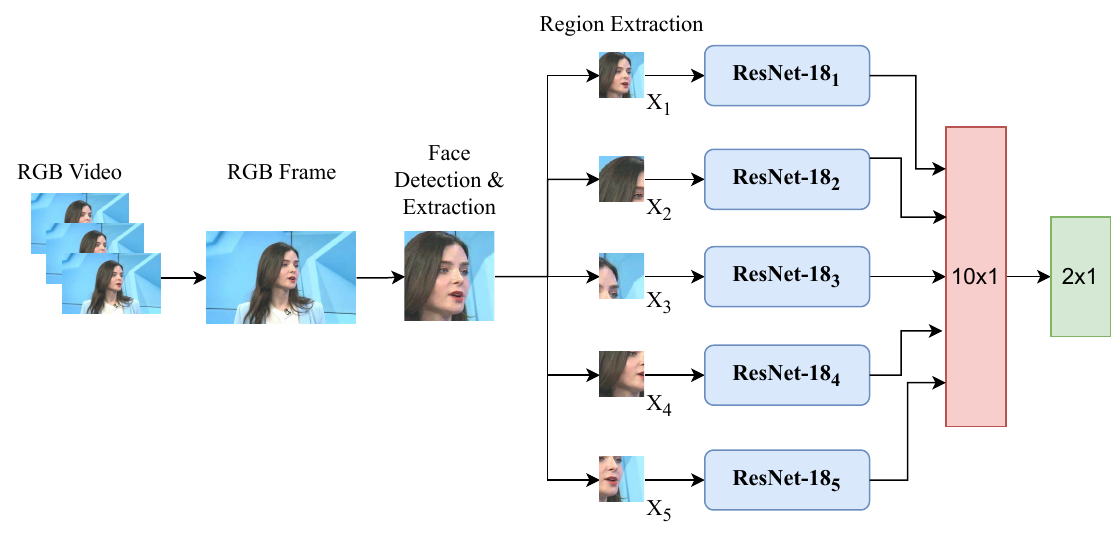}
    \caption{Ensemble learning approach of \cite{kumar2020detecting} which incorporates numerous state-of-the-art classification models.}
    \label{fig:ensemble_model}
\end{figure}

Ensemble learning approaches have also been leveraged in literature. For instance, Kumar et al. \cite{kumar2020detecting} proposed to use five ResNet18 models to extract local and global features. Specifically, one ResNet architecture learns overall facial attributes and the remaining four
are dedicated to learning the local, regional attributes. The outputs from these five parallel ResNet-18s, which represent the classifications from the models only considering their respective inputs are concatenated to form a 10-dimensional vector. Then the weighted fusion of these individual scores is performed to generate the final binary classification. This architecture is visually illustrated in Fig. \ref{fig:ensemble_model}.

When training this framework the authors have utilised a series of cross-entropy losses. The total loss is composed of the sum of the cross-entropy loss of the full-face model, the entropy loss of local regional models, and the cross-entropy loss after the final fusion.  

Another ensemble learning architecture is in \cite{rana2020deepfakestack} in which the authors propose to leverage seven state-of-the-art deep learning models to extract representations. Specifically, XceptionNet, MobileNet, ResNet101, InceptionV3, DensNet121, InceptionReseNetV2, and DenseNet169 models initialised with ImageNet weights have been used as base learners. The authors have replaced the last two layers with a layer with softmax activation. Greedy Layer-wise Pretraining is used to finetune these base learners. Given an input image, these base learners generate true or fake class predictions and the authors propose a stack generalisation model which learns to pick the best combination of the prediction considering the outputs of individual base learners. This framework is trained using the categorical cross-entropy loss. 

In \cite{sabir2019recurrent} the authors argue that less attention has been paid to the temporal features for the detection of deepfakes and propose a recurrent convolutional framework. In their proposed approach the first step is to detect, crop, and align the faces to a reference coordinate system such that any rigid motion of the face is compensated. In the next stage face manipulation detection is conducted using a recurrent convolutional network where the encoding of the frame-wise features is done using the CNN backbone and the final prediction is generated by a recurrent neural network via analysing those sequences of features. Both ResNet and DenseNet have been experimented as the backbone architecture. Moreover, the authors propose to extract features at multiple levels from the backbone CNNs, and these features are processed by individual recurrent networks. This framework is trained end-to-end using cross-entropy loss for binary classification. 

More recently, the success of transformer networks in modelling spatiotemporal features has seeped into the face deepfakes detection domain. In \cite{wodajo2021deepfake} a Convolutional Vision Transformer model is proposed for the detection of deepfakes. Specifically, the CNN architecture is capable of extracting discriminative features from the individual frames and the transformer module learns to analyse the correlation across the sequence of these features and classify them
using an attention mechanism. The authors named the feature extraction CNN as the Feature Learning (FL) component which is a stack of 17 convolutional blocks. The transformer block which receives the feature map of FL is identical to the ViT architecture in \cite{dosovitskiy2020image}. This framework is trained using the binary cross-entropy loss function. In a similar line of work \cite{zhao2023istvt} proposes an Interpretable Spatial-Temporal Video Transformer (ISTVT) for deepfake detection. This architecture leverages a feature extractor constructed using Xception blocks to extract salient textures from the input face. These feature maps are decomposed into tokens. spatial and temporal self-attention modules are used to attend to both dimensions. Specifically, in temporal self-attention the attention heads attend to patches of the same location across the frames while in spatial attention all the patches in each frame are considered. This decomposition of self-attention enabled the authors to interpret the model across both dimensions. ISTVT framework is trained by the binary cross-entropy (BCE) classification loss.

In \cite{xu2022supervised} the authors motivate the need for the deepfake detection algorithm to be agnostic across generation type, quality, and appearance. Furthermore, they argue the need for contrastive learning as the appearance characteristics of the fake video is highly indistinguishable. The authors first train an encoder network which learns to generate normalised embedding from augmented data.  A projection network then uses these embeddings and computes the supervised contrastive loss. Finally, a linear classifier is trained using cross-entropy loss to discriminate between real and fake faces. In another work \cite{dong2023contrastive} multiple views of the same image are used as the augmentation for contrastive learning. The authors observe that the deeper feature representation tends to focus on semantic information while the artifacts left by deepfake generation algorithms exist in shallow feature maps and propose a multi-scale feature enhancement module to combine both local and global features. Furthermore, the authors argue that the deepfake generation artifacts can be found in the frequency feature domain and propose a Steganalysis Rich Model (SRM) \cite{fridrich2012rich} to extract local noise features from neighboring pixels. The authors propose a combination of cross-entropy loss and consistency loss to train this framework. Specifically, the consistency loss minimises the cosine distance in feature space for different augmentations of the same image and cross-entropy loss supports the deepfake detection. 

When considering the multimodal approaches for face deepfake detection, \cite{mittal2020emotions} and \cite{zhou2021joint} are notable considering their effectiveness. In \cite{mittal2020emotions} a two-branch architecture is adopted to process features from both real and fake videos. Specifically, the authors propose to extract facial features using OpenFace \cite{baltruvsaitis2016openface} and speech features using pyAudioAnalysis \cite{giannakopoulos2015pyaudioanalysis} from the raw videos. The extracted features are passed through the two-branch neural network architecture where a separate branch is used for processing each modality separately. Within each branch, there are two separate networks for extracting features representing the modalities and perceived emotions. These feature vectors, each with 250-dimensions, is used to compute a triplet loss function that  minimise the similarity between the modalities from the fake video and maximise the similarity between modalities for the real video. Another multi-stream network architecture is in \cite{zhou2021joint} which uses audio and video streams for the detection of deepfakes. To fuse the video and audio streams the authors propose to apply central connections. At each layer, the audio and visual representation will be fused with the current layer of sync-stream and used as input to the fusion at the next layer. This is achieve through, (i) inter-attention: which computes attention across visual and audio representations, (ii) Inter+intra-attention: which is the video or audio modality-specific self-attention, and (iii) Joint-attention: where the authors have applied same attention weights on both visual and audio representations. During the inference stage, preliminary predictions are obtained through the sync-stream and if it is a positive prediction video and audio branches will be individually analysed to generate the final prediction.

\noindent\textbf{Anomaly detection-based approaches: } In contrast to the deep learning-based approaches which learn discriminative features that could differentiate real faces from fake ones, the anomaly detection-based approaches are designed to learn the distribution of real faces. An input face that significantly deviates from the learned real distribution is identified as a fake face. 

\begin{figure}
    \centering
    \includegraphics[width=.6\linewidth]{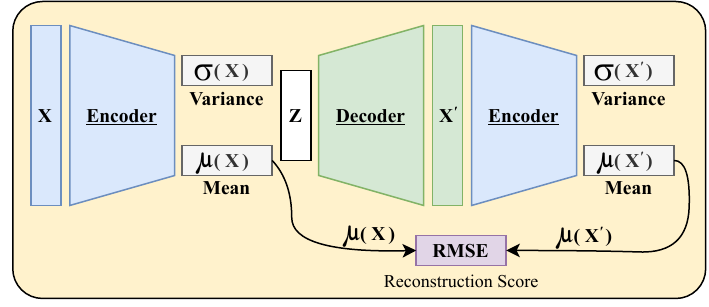}
    \caption{OC-FakeDect-2 architecture proposed in \cite{khalid2020oc}which is based on VAE and detects which compares the reconstruction error of the input image and the reconstructed image in the latent space for deepfake detection.}
    \label{fig:OC-FakeDect-2}
\end{figure}

In \cite{khalid2020oc} the authors propose the OC-FakeDect framework which is formulated as a VAE-based approach. Two versions of the OC-FakeDect network architecture are proposed. In the first version of the model the authors propose to compare the input and reconstructed images directly in the image space using Root Mean Square Error (RMSE). In the next version, an additional encoder is appended to map the reconstructed image back to the latent space and the authors propose to compare the input and reconstructed images in the latent space. This architecture is visually illustrated in Fig. \ref{fig:OC-FakeDect-2}. For training this framework the authors have used the KL divergence loss to force the network to approximate a Gaussian distribution in the latent space and mean square error to help minimise the error between the input and reconstructed images. 

A two-step approach that learns the probability of an occurrence of a certain pixel based on its neighbourhood is proposed in \cite{khodabakhsh2020generalizable}. The authors condition each pixel on pixels before (in raster order) and extend the PixelRNN model \cite{van2016pixel} to learn this distribution for real images. This learned model, named PixelCNN++, predicts the probability distribution that denotes the likelihood of observing a specific pixel value at a given location considering all pixel values before it. Using this approach the authors propose to calculate a probability matrix for the entire image which denotes the likelihood of observing the input image. A Universal Background Model (UBM) is trained with the PixelCNN++ to further refine the features.  A simple classifier based on LeNet-5 \cite{lecun1998gradient} is trained on the output of the UBM model to generate the real/fake classification. 

We would also like to note the contrastive learning approach proposed in \cite{cozzolino2023audio} that exploits the audio-visual features that are exhibited in real videos. Specifically, they propose to extract audio and face embedding vectors, and at each training iteration, these features from $N$ input videos are extracted. By comparing only-video, only-audio, and audio-video feature vectors of $N$ inputs three $N \times N$ similarity matrices are computed. The authors have then utilsed three contrastive losses, one for each similarity matrix, to push the embedded vectors of the same individual closer and move those of different individuals farther apart. For training this framework the overall loss is defined by aggregating the three contrastive losses. During the test time, the authors assume that they have at least 10 real videos of the person of interest and calculate a similarity matrix between the features of the test video and this set of reference videos. the mean and standard deviation of the similarity index are calculated which is used to make a decision regarding the authenticity of the test video. 

\noindent\textbf{Summary of face deepfake detection methods and open research questions:} When reviewing the literature it could be seen that variety of features have been proposed to date for the task of face deepfakes detection. They range from blinking patterns, biological signals such as PPG signals, and 3D head poses to facial behavioral features. Despite these advances to date, there is no universal face deepfake detection methodology that could withstand the current and future advances of the face deepfake generation technology. For instance, most of the existing state-of-the-art face deepfakes detection methods such as \cite{mittal2020emotions} and \cite{zhou2021joint} are not robust against external face deepfake generation methodologies that haven't been seen during the training. Furthermore, they poorly generalise to unseen datasets. As such, the current research achievements within face deepfake detection are far from producing a universal face deepfake detector and warrant further research efforts. Moreover, the lack of interpretability of the face deepfake detection methods has been a major limitation in order to build trust in the general public regarding their decisions. In Tab. 3 of supplementary material we provide a summary of different face deepfake detection methods, highlighting their strengths and weaknesses.
\subsubsection{Performance evaluation of deepfake detection methods}

For evaluating the efficacy of the face deepfake detection algorithms several different metrics have been used in the literature. Among them, Accuracy, Precision, Recall, F1-Score, Area Under the ROC Curve (AUC), and Error Rate are the most commonly used. For details please refer to the supplementary material section 4.4 on ``Introducing standard evaluation protocols''.

\begin{table*}[htbp]
\caption{Top 10 Tools to Detect Face Deepfakes}
 \resizebox{\textwidth}{!}{%
\begin{tabular}{|c|cccc|c|c|c|}
\hline
\multirow{2}{*}{Method}          & \multicolumn{4}{c|}{Type of Deepfake that it Can Detect}                                          & \multirow{2}{*}{Free} & \multirow{2}{*}{Open-source} & \multirow{2}{*}{URL}                              \\ \cline{2-5}
                                 & \multicolumn{1}{c|}{Image} & \multicolumn{1}{c|}{Audio} & \multicolumn{1}{c|}{Video} & Multimodal &                       &                              &                                                   \\ \hline
Sentinel                         & \multicolumn{1}{c|}{\cmark}     & \multicolumn{1}{c|}{\cmark}     & \multicolumn{1}{c|}{\cmark}     & \cmark          & \xmark                     & \xmark                            & https://thesentinel.ai/                           \\ \hline
Sensity                          & \multicolumn{1}{c|}{\cmark}     & \multicolumn{1}{c|}{\cmark}     & \multicolumn{1}{c|}{\cmark}     & \cmark          & \xmark                     & \xmark                            & https://sensity.ai/                               \\ \hline
Microsoft Video AI Authenticator & \multicolumn{1}{c|}{\cmark}     & \multicolumn{1}{c|}{\xmark }     & \multicolumn{1}{c|}{\cmark}     & \xmark          & \cmark                     & \xmark                            & https://blogs.microsoft.com                       \\ \hline
Deepware                         & \multicolumn{1}{c|}{\cmark}     & \multicolumn{1}{c|}{\xmark }     & \multicolumn{1}{c|}{\cmark}     & \xmark          & \cmark                     & \xmark                            & https://deepware.ai/                              \\ \hline
Intel's FakeCatcher              & \multicolumn{1}{c|}{\cmark}     & \multicolumn{1}{c|}{\xmark }     & \multicolumn{1}{c|}{\cmark}     & \xmark          & \cmark                     & \xmark                            & https://www.intel.com                             \\ \hline
DeepReal                         & \multicolumn{1}{c|}{\cmark}     & \multicolumn{1}{c|}{\xmark }     & \multicolumn{1}{c|}{\cmark}     & \xmark          & \cmark                     & \xmark                            & https://deepfakes.real-ai.cn/                     \\ \hline
CADDM                            & \multicolumn{1}{c|}{\cmark}     & \multicolumn{1}{c|}{\xmark }     & \multicolumn{1}{c|}{\cmark}     & \xmark          & \cmark                     & \cmark                            & https://github.com/megvii-research/CADDM          \\ \hline
ID-Reveal                        & \multicolumn{1}{c|}{\xmark }     & \multicolumn{1}{c|}{\xmark }     & \multicolumn{1}{c|}{\cmark}     & \xmark          & \cmark                     & \cmark                            & https://github.com/grip-unina/id-reveal           \\ \hline
Audio Visual Forensics           & \multicolumn{1}{c|}{\cmark}     & \multicolumn{1}{c|}{\cmark}     & \multicolumn{1}{c|}{\cmark}     & \cmark          & \cmark                     & \cmark                            & https://github.com/cfeng16/audio-visual-forensics \\ \hline
DuckDuckGoose                    & \multicolumn{1}{c|}{\cmark}     & \multicolumn{1}{c|}{\xmark }     & \multicolumn{1}{c|}{\cmark}     & \xmark          & \xmark                     & \xmark                            & https://www.duckduckgoose.ai/                     \\ \hline
\end{tabular}}
\label{tab:tool_det_face_deepfakes}
\end{table*}

\noindent\textbf{Top-ten tools for detecting face deepfakes:} Tab. \ref{tab:tool_det_face_deepfakes} summarises top-ten tools, including free and open-source tools, that can be leveraged for the detection of face deepfakes. It should be noted that for the ranking of these methods, we consider the accuracy of the detection, efficiency, the modalities that the detection algorithm can consider, and the availability of the source codes.

\hspace{2mm}
\subsection{Combating face deepfakes in face biometrics}
While it is difficult to fool physical biometric authentication systems using face deepfakes, online authentication systems such as mobile-based personal authentication systems can be fooled using state-of-the-art deepfake technology as they can counter liveliness detection method such as micro-muscle movements, eye-blinking patterns. For instance, a recent Gartner report \footnote{https://www.gartner.com/en/newsroom/press-releases/2024-02-01-gartner-predicts-30-percent-of-enterprises-will-consider-identity-verification-and-authentication-solutions-unreliable-in-isolation-due-to-deepfakes-by-2026} predicts that by 2026, due to face deepfakes, 30\% of enterprises will no longer be able to consider face biometric and authentication solutions to be reliable in isolation.
As such, it is important to investigate the ability of off-the-shelf face deepfakes to thwart state-of-the-art biometric recognition models. 

The following subsection provides a summary of the results of this investigation and we refer the reader to Sec. 3 of supplementary material for detailed comparisons.

\subsubsection{Summary of the efficacy of face deepfakes to fool face biometrics systems}
Evaluation results of face deepfakes on face biometrics are presented in Tab. 4 of the supplementary material and provide alarming evidence demonstrating the ability of both face swap and face reenactment methods to thwart state-of-the-art face recognition models. Our evaluations demonstrate that deepfake methods such as Wav2Lip, SimSwap, and first-order model are capable of fooling biometric recognition systems, especially the lightweight systems such as MobileNet. This vulnerability is of significant concern due to the vast utilization of lightweight face verification methods for authentication in applications such as mobile device unlocking, app login and payment gateways, and in social media apps for photo tagging.

\subsubsection{Measures for revealing true identity: } While significant research has been conducted in the area of deepface detection, solutions for reversing a faked face, resulting from manipulations, to recover the original real face,  are yet to be developed. \textbf{To date there has been only a single work on this topic.} This framework, introduced by Chang et. al \cite{chang2023cyber}, works using a pair of neural network modules, named, vaccinator and neutraliser for manipulation reversal. Within their conceptual framework the deepfake generator sits in between the vaccinator and neutraliser and these two models jointly attack the model in the middle. Specifically, the vacinator learns to synthesize the face region of an original image based on the mask and the neutraliser leverages this mask and the deepfake image in which the face region has been masked to reconstruct the original face. Due to the joint training of both vaccinator and the neutraliser, during the vaccination stage the vacinator injects identity-specific features which the neutraliser could leverage to recover the true identity. We discuss the area of revealing true identity of a manipulated face in Section 4 on Future Research Directions in the supplementary material.
\section{Applications of deepfakes} \label{sec:applications}
In this section, we present a spectrum of positive applications of deepfakes ranging from fashion to the entertainment industry. Moreover, the ethical, psychological, and security implications of deepfakes are also discussed in this section. 

\subsection{Positive commercial applications}

Despite its controversial reputation, the deepfake technology holds great potential for positive commercial applications across a myriad of industries. In addition to the reenactment of passed actors with hyper-realistic visual quality, which we see being readily used in the entertainment industry, there are various current and future use cases of deepfake technology across a diverse set of applicational areas if this technology is used ethically and responsibly. We use this section to introduce such sample use cases.  

\subsubsection{Fashion and beauty industry:}
With the rise of deepfake technology, the way that leading brands engage with consumers has been revolutionised. In 2021 Kati Chitrakorn, the Vogue business technology expert predicted that deepfake technology would transform the landscape of the fashion industry \cite{deepfakevoguebusiness}. 

For instance, \textbf{digital fashion shows and influencer collaborations} has now become a reality \cite{deepfakeglamour}. The onset of the COVID-19 pandemic only accelerated the use of deepfake technology enabling the collaborations and interactions between people in a setting where in-person activities are restricted. Specifically, Demna Gvasalia's `deepfake' Spring 2022 Balenciaga fashion show illustrated how deepfake technology can be used to model garments by celebrities or influencers with just a few sets of images without the need to even be there in person.  

In the 2019 London Fashion Week, some selected sets of participants were able to watch themselves wearing HANGER’s latest collection using deepfake technology. These \textbf{virtual try-ons} were being projected behind the live models enabling the members of the audience to see their appearance if they were actually wearing the garments that were being modelled. Superpersonal \footnote{https://www.producthunt.com/products/superpersonal} is an app that has been specifically designed to allow users to try on clothes virtually. This allows the consumers to visualize products before purchase and understand which styles fit their taste better. Several other innovations have also emerged in this direction. Fxgear's FXMirror \footnote{https://www.fxgear.net/vr-fashion} which provides an augmented-reality fitting room experience such that the shoppers can try on clothes virtually and YourFit solution \footnote{https://3dlook.ai/yourfit/} by 3DLOOK are only a few examples. 

\subsubsection{Marketing industry}
 
Furthermore, the flexibility and customisability of the synthesised media have no limits. As such \textbf{hyper-personalised advertisements} could be generated targeting consumers in different demographics by customising the clothing, voice, and location of characters in the advertisement. Unethical and fraudulent use of this technology was experienced in Taylor Swift's Deepfake Campaign where her image was used without permission for the creation of advertising media that promoted products she did not endorse. However, this technology can be ethically used with the permission of the celebrity or the influencer to create highly influential advertisements that could reach a far greater audience with minimal cost. The 2019 malaria awareness deepfake advertisement that featured David Beckham \footnote{https://youtu.be/QiiSAvKJIHo?si=RuNnN5hE1R78JTCQ} is a prime example that demonstrates this marketing potential. In this video, David Beckham speaks in nine different languages appealing to end malaria and multimodal deepfake technology has made Beckham appear multilingual. German online retail giant Zalando has also been readily using deepfake technology in its marketing campaigns. For example, the deepfake technology enabled supermodel Cara Delevingne to appear in 290,000 localised advertisements \cite{deepfakevoguebusiness}. deepfake technology in marketing can be further extended to reenact historical figures and bring them to life with contemporary public figures. This will \textbf{enhance the storytelling} of the marketing materials and better capture the attention of the targetted consumers. 

The deepfake technology is transforming the marketing industry as a whole. The deepfake technology reduces production costs by cutting down the costs associated with the hiring of the production crews, including, videographers, camera operators, media editors, casting assistants, and directors. Furthermore, it doesn't need a location to shoot the videos or equipment to record them. Hour One \footnote{https://hourone.ai/} is a company that readily uses deepfakes to create commercial media. In one of their advertisement campaigns human actors are replaced by animated digital clones of real humans generated by deepfakes \cite{deepfaketechnologyreview}. The impracticality of using real actors and production crews to create thousands of videos in different languages has led Hour One to opt for deepfakes.

\subsubsection{Corporate training, simulation, and virtual assistants}

The British multinational company WPP has created training videos together with Synthesia, a synthetic media generation company, targeting its employees \cite{deepfakewired}. These videos have been sent to thousands of employees that WPP has worldwide and address the employee by name and explain some basic concepts in AI. This example shows how deepfake technology can be used to create \textbf{corporate training} material in a personalised and cost-effective manner, which is otherwise impractical for a multinational company. However, the merits of deepfake technology in corporate training and simulation settings surpass this simple application. For instance, it can be used to create highly realistic customer interactions or crisis management situations for training purposes. The AI models are adaptable and they can dynamically change their responses based on the responses of the trainee, providing a personalised training experience. Moreover, the deepfake technology provides a risk-free training environment for areas such as law enforcement and healthcare, therefore, trainees can practice decision-making in critical settings.  

The Live Interactive Customer Experience (ALICE) receptionist \footnote{https://www.alicereceptionist.com/} is an ideal example of how \textbf{virtual receptionist kiosk} can be used to handle visitors' queries, replacing the role of a human receptionist. We incorporate the virtual receptionist application under the deepfake category considering the use of deepfake technology to create human-like avatars that represent the virtual receptionist. Furthermore, this virtual assistant is capable of conducting a range of visitor management tasks, including, pre-visit check-ins, visitor screening, driver's license scanning, and body temperature check. These virtual receptionist kiosks are currently being used in various American International Group and ING Group branches.

\subsubsection{ Entertainment industry: }

deepfake technology is widely being used for \textbf{dubbing or revoicing media in the entertainment industry}. This allows the synchronise facial expressions, lip movements, and expression of emotions after the dubbing process. An AI-driven startup company named Flawless \footnote{https://www.flawlessai.com/} is generating deepfake dubs which is cost-effective and efficient, and help the media content reach new audiences. Compared to traditional dubbing methods which have mistimed mouth movement, Flawless utilises deepfake technology to artificially synthesise lip movements that match the translated speech. The result is a much smoother revoicing of the media. 

The movie industry is heavily utilising Computer-Generated Imagery (CGI) to create visual effects. This is a meticulous process done by visual effects artists. The deepfake technology has the potential to automate this process by generating different renders of the chosen character automatically. The gaming industry has started adapting the deepfake technology. For instance, in the video game Cyberpunk 2077 \footnote{https://www.cyberpunk.net/au/en/} where celebrities play roles in the game. We believe this technology will soon seep into the film industry as well giving more potential to the filmmakers, enabling scenarios such as reenacting historical events or bringing characters to life.  

\hspace{2mm}

\subsection{Negative implications}
This subsection summarises the primary negative implications of deepfake technology and highlights the need for the urgent need for countermeasures. 

\subsubsection{Misinformation and disinformation propaganda}

In mid-March 2022 a deepfake video of Ukrainian President Volodymyr Zelenskyy appeared on social media calling on the Ukrainians to stop fighting and to surrender their weapons \cite{deepfakenortheastern}. This video was even broadcast on the Ukrainian television channel Ukraine 24 by a team of hackers. 

The harmful effects of mis/disinformation generated through deepfakes do not stop from fake news. It can be used to impact elections, perform corporate sabotage with well-times and articulated falsifying evidence, and harm the image of public figures. Most importantly these mis/disinformation campaigns could deteriorate public trust regarding the authenticity of genuine material in mainstream media. For instance, when the Princess of Wales, released a video statement in March 2024 sharing that she had been diagnosed with cancer a fresh round of conspiracies regarding deepfakes reappeared in social media \cite{deepfakewashingtonpost}. However, this time it was people disbelieving a real video. These real-world examples clearly elaborate the growing threat of multimodal deepfake to society in an era where seeing is not believing.

Apart from these recent examples, there are other evidence of deepfake being used to spread mis/disinformation. It has been suggested that a deepfake story could have sparked the diplomatic confrontation between Saudi Arabia and Qatar \cite{deepfakebrookings1}. The unnatural speech of President Ali Bongo sparked a military coup in 2018 in Gabon claiming that the video was a deepfake and the president was no longer healthy enough for the office or even had died \cite{deepfakeforbes}. Furthermore, there was an unsuccessful attempt to discredit and overthrow Malaysia’s economic affairs minister using deepfake-based fabricated media \cite{deepfakewii}. 

Moreover, deepfakes can be used to create fake influences or endorsements. For instance, fraudulent Taylor Swift advertisements that promoted a cookware brand on social media are a prime example of such fraudulent narratives. Based on our review of deepfake detection methods, there is no universal deepfake detector that could suffice and withstand all the advances of current and future deepfake generation technology, and until such robustness is met our society faces ongoing threats due to the malicious use of deepfake technology.

\subsubsection{Psychological impact}
The psychological impact of deepfakes is quite concerning as it affects not only individuals on a personal level but also at social and societal levels. 

Deepfakes disrupt our ability to believe what we perceive. Therefore, it could lead to deterioration of trust of people regarding news and media in general. This effect occurs even if the deepfake is unsuccessful in misleading a particular individual. The sense of deception leads to increased skepticism and uncertainty in our daily online and offline interactions. For instance, a study by Vaccari and Chadwick \cite{vaccari2020deepfakes} found that even if a person is not completely misled by a deepfake the exposure to it reduces their trust in news.

Another study found alarming evidence of deepfakes modifying our memories and even implanting false memories. For instance, in \cite{dobber2021microtargeted} the authors found that watching deepfake videos could result in participants falsely remembering nonexistent films. Furthermore, it could lead to a change in one's attitude. Specifically, the authors of \cite{dobber2021microtargeted} constructed multimodal deepfakes and have shown them to a selected group of individuals to see if there is any change in their attitudes toward the politician and the attitudes toward his or her political party. This study revealed that microtargeting the deepfake to groups that are most likely to be offended could amplify its negative implications.   Furthermore, we should consider the profound emotional impact of deepfakes on the individuals whose identities have been maliciously depicted in the video. The fabricated media could be embarrassing, offensive, and damaging to their reputation, leading to anxiety, and even altering their beliefs and behaviour.

\subsubsection{National security threat of deepfakes} 

Lt. Gen. Jack Weinstein who is the deputy chief of staff for strategic deterrence and nuclear integration at the United States (US) Air Force Pentagon headquarters stated ``The greatest existential threat to the United States of America is the fracturing of our democracy and the intentional misleading of facts to support political agendas''. Therefore, it is clear that false or misleading information that is deliberately spread to deceive a population could cripple the world's largest economy and the second-largest democracy. Furthermore, based on the press release of The US National Security Agency on September 12, 2023 \cite{deepfakeNSA}, synthetic media can cause public unrest through the spread of false information about political, social, military, or economic issues. This report states that public availability of the implementation of deepfake generation algorithm has made mass production of fake media easier and less expensive, which has broadened their impact to a larger scale.  

The national security threat of deepfakes also includes cyber espionage through impersonation where deepfake technology can be used to fabricate fake communications of high-ranking officials. While to the best of our knowledge, this has not occurred to date, there exists evidence to the use of audio deepfakes has been used to steal personally identifiable information during fake online interviews of potential applicants \cite{deepfakeFBI}. This information can be used to create fake credentials to gain access to sensitive information or critical infrastructure systems. 

Moreover, one should consider the economic impact of deepfake as the economy is closely associated with national security. This is clearly evident by the findings in the 2024 global risk report of the  World Economic Forum. This report states that misinformation and disinformation are the biggest short-term risks to the world economy. For instance, meticulously targeted false information about large corporations of a certain country could be used to disrupt markets or manipulate stock prices leading to economic instability in that particular country. 

Government agencies, researchers, and policymakers should continue to collaborate together to minimise the threat of deepfakes to national and global security. Furthermore, mainstream media has a major role in promoting the awareness and literacy of the general public regarding the threat of deepfakes and how to spot them.

\subsubsection{ Privacy violation} 
A popular example of deepfakes in privacy violation is the creation and distribution of pornographic material by swapping an individual's face, voice, and body into real pornography. For instance, the Reddit user made deepfake sex videos of female celebrities using their images and videos. However, the potential victims are not limited to public figures. It can be used to generate revenue or targeted attacks against specific individuals, such as ex-partners, or rivals which is an invasion of sexual privacy \cite{citron2018sexual}.

The impact of such activities is not limited to emotional distress, reputational damage, and personal trauma. Blackmailers could use deepfakes to extortion. The victims may be forced to provide money or even business secrets to prevent the release of the deepfakes. Several legislative changes have been proposed to protect victims from privacy-related issues caused by deepfakes, however, it has been identified that there exist several issues when dealing with deepfakes in litigation and governments need to take more actions to protect victims \cite{deepprivacy}.

\hspace{3mm}
\section{Future research directions} \label{sec:future_research}
 
We refer the readers to Sec. 4 in the supplementary material, where we discuss future research directions, including the design of universal deepfake detection methods, recovering the true identity, explainable deepfakes detection methods, introducing standard evaluation protocols, and design of a regulatory framework for the governance of deepfake research. 

\section{Conclusion}\label{sec:conclusions}
In this survey paper, we have discussed existing state-of-the-art methods for the generation and detection of face deepfakes. Our analysis emphasises an algorithmic perspective, providing an in-depth discussion of the architectures, and include details such as training paradigms, loss functions and evaluation metics. 
In addition, we have discussed the biometric implications of the generated face deepfakes and we have provided in-depth discussion regarding their positive and negative applications. As concluding remarks, we outlined key research gaps and proposed possible future research directions for further investigation. 
\appendix
\section{Summary of state-of-the-art face swap deepfake generation approaches}
In Tabs. \ref{tab:face_swap_summary} and \ref{tab:face_reenact_summary} we summarise the state-of-the-art face-swap and face-reenact deepfake generation methodologies, respectively, and discuss their strengths and weaknesses. 

\begin{table*}[htbp]
\caption{Summary of state-of-the-art face swap deepfake generation approaches. NA indicates that quantitative comparisons are not available.}
\resizebox{\textwidth}{!}{%
\begin{tabular}{|p{2cm}|p{2cm}|p{2cm}|p{2cm}|p{5cm}|p{5cm}|}
\hline
Method                  & Input Features                                                 & Architecture                    & Performance & Strengths                                                                                                           & Weaknesses                                                                                                                                                                                      \\ \hline
FCN \cite{nirkin2018face}           & Facial images and landmarks                                    & CNN                             &      FaceForensics++ \cite{rossler2019faceforensics++} 42.1 (LMD), 0.45  (ID)         & Introduces a semi-supervised pipeline for training the face segmentation model                                      & Requires extensive training data to train and the quality of the face swapping depends on the training data.                                                                                    \\ \hline
Face-swap GAN \cite{deepswapgan}  & Facial images and landmarks                                    & GAN                             &      NA       & Simple encoder-decoder-discriminator architecture. Reasonable handling of occlusion. Generate realistic eye regions & Can only swap faces between specific identities. Generates overly smoothed faces, face alignment is required.                                                                                   \\ \hline
DeepFaceLab \cite{perov2020deepfacelab}   & Facial images and landmarks                                    & GAN                             &    FaceForensics++ \cite{rossler2019faceforensics++}  0.73 (LMD)         & High resolution generations. Mature open access toolkit                                                             & The quality of the synthesisation is tightly coupled with the face segmentation quality.                                                                                                        \\ \hline
FSNet \cite{natsume2019fsnet}        & Facial images and landmarks                                    & VAE-based encoder-decode +  GAN &    FaceForensics++ \cite{rossler2019faceforensics++} 30.8 (LMD), 0.36 (ID)          & Subject-agnostic                                                                                                    & Cannot handle occlusions. Synthesised faces have poor resolution compared to resent state-of-the-art methods                                                                                    \\ \hline
RSGAN \cite{natsume2018rsgan}         & Facial images and landmarks                                    & Facial Region separator + GAN   &  CelebA \cite{liu2015deep} 1.127 (AED)         & Subject-agnostic. Can also be used to edit facial attributes                                                        & Synthesised faces have poor resolution compared to resent state-of-the-art methods                                                                                                              \\ \hline
FS-GANv2 \cite{nirkin2022fsganv2}     & Facial images and landmarks                                    & GAN                             &    FaceForensics++ \cite{rossler2019faceforensics++} 21.6 (LMD), 0.37 (ID)          & Can handle occluded faces. Capable of face swapping and face reenactment.                                           & Reliant on facial landmark detection which can sometimes produce erroneous landmarks. Iterative architecture that uses only one frame at a time which does not utilise any temporal information \\ \hline

Faceshifter \cite{li2019faceshifter}   & Facial images and facial attributes extracted from these faces & GAN                             &    FaceForensics++ \cite{rossler2019faceforensics++}  45.51 (LMD), 0.60 (ID)           & Visually appealing results with consistency in pose, expression, and lighting.                                      & Cannot generate high-resolution images. Iterative frame-by-frame processing which does not incorporate temporal information                                                                     \\ \hline
SimSwap \cite{chen2020simswap}       & Facial images                                                  & GAN                             &    FaceForensics++ \cite{rossler2019faceforensics++} 8.04 (EFD), 11.76 (FID)           & Effective injection of source identity,                                                                             & Cannot handle occlusions. Limited resolution in the synthesised faces                                                                                                                           \\ \hline
HiRFS \cite{xu2022high}         & Facial images and landmarks                                    & GAN                             &  FaceForensics++ \cite{rossler2019faceforensics++} 2.79 (EFD)           & Disentanglement of semantics within the latent space. Introduces specialised losses to enable temporal cohenerency. & The quality of the synthesised results depends on the latent codes produced by the  StyleGAN model. Therefore, not guaranteed to preserve the identity attributes of the source face            \\ \hline
MegaFS \cite{zhu2021one}       & Facial images                                                  & GAN                             &   FaceForensics++ \cite{rossler2019faceforensics++} 2.96 (EFD)          & High resolution face swapping. Can manipulate multiple latent codes concurrently.                                   & The quality of the synthesised results depends on the latent codes produced by the  StyleGAN model                                                                                              \\ \hline
FaceDancer \cite{rosberg2023facedancer}    & Facial images                                                  & GAN                             &    FaceForensics++ \cite{rossler2019faceforensics++} 7.97 (EFD), 16.30 (FID)          & Better preservation of facial attributes of the source face.                                                        & Limited resolution in the synthesised faces. Limited robustness in occlusions and in poor lighting conditions.                                                                                 \\ \hline
\end{tabular}}
\label{tab:face_swap_summary}
\end{table*}

\begin{table*}[htbp]
\caption{Summary of state-of-the-art face reenactment deepfake generation approaches. NA indicates that quantitative comparisons are not available. }
 \resizebox{\textwidth}{!}{%
\begin{tabular}{|p{2cm}|p{5cm}|p{2cm}|p{2cm}|p{5cm}|p{5cm}|}
\hline
Method                    & Input Features                         & Architecture & Performance & Strengths & Weaknesses \\ \hline
Face2Face \cite{thies2016face2face}       &    Facial images and landmarks                                    &      3D Morphable Face Models         &           NA  &   A semi-supervised architecture        & Cannot handle occlusions and different head poses           \\ \hline
ReenactGAN \cite{wu2018reenactgan}      &        Facial images                                 &    GAN          &   DISFA \cite{mavadati2013disfa} 58.4 \% (Facial Action Units Accuracy)       &   robust to different poses, expressions and lighting conditions        &     Output resolution is poor       \\ \hline
GANimation \cite{pumarola2018ganimation}      & Facial images and emotion action units &        GAN      &       NA      & Can handle complex backgrounds and illumination conditions          &     Unable to adapt to gaze variations       \\ \hline
FOMM \cite{siarohin2019first}            &    Sparse key-points                                    &   GAN           &  VoxCeleb2 \cite{nagrani2020voxceleb}   0.043 (L1)       &    Can handle complex motions and can animate diverse object types       &    Complexities when handling dynamic backgrounds        \\ \hline
Talking Heads \cite{zakharov2019few} &    Facial images and landmarks                                    &          GAN    &     VoxCeleb2 \cite{nagrani2020voxceleb} 30.6 (FID)        &   A few-shot learning architecture        &   Cannot manipulate gaze         \\ \hline
FC-TFG \cite{jang2023s}                 &    Facial images and audio                                    &           GAN   &    VoxCeleb2 \cite{nagrani2020voxceleb} 1.58 (LMD)         &  controllable head pose, eyebrows, eye blinks, eye gaze, and lip movements         &  Requires multimodal inputs          \\ \hline
Multimodal Talking Faces \cite{yu2020multimodal}                &   Source audio and target face video                                     &  GAN            &      In house Trump Dataset 0.889 (SSIM)       &     Audio driven reenactment      &  Only generates faces with limited pose variations          \\ \hline
EmoGen \cite{goyal2023emotionally}                 &        Audio, video and emotions                                &    GAN          &    CREMA-D \cite{cao2014crema} 6.04 (FID)         &  can generate faces with diverse emotions         &    has been evaluated with straight head poses        \\ \hline
AVFR-GAN \cite{agarwal2023audio}        &   Facial images and audio                                     &     GAN         &   VoxCeleb \cite{nagrani2017voxceleb} 8.48  (FID)        &  Generalises well to unseen faces         &     cannot handle occlusions       \\ \hline
PNCC GAN  \cite{xue2023high}                &    3D face                                    & GAN             &    VoxCeleb \cite{nagrani2017voxceleb} 17.21 (FID)         &       preserves target face identity    &  cannot handle extreme head poses          \\ \hline
\end{tabular}}
\label{tab:face_reenact_summary}
\end{table*}

\section{Summary of face deepfake detection methods}

In Tab. \ref{tab:detection_summary} we provide a summary of different face deepfake detection methods, highlighting their strengths and weaknesses.

\begin{table*}[htbp]
\caption{Summary of face deepfake detection approaches}
 \resizebox{\textwidth}{!}{%
\begin{tabular}{|p{2cm}|p{1cm}|p{5cm}|p{3cm}|p{5cm}|p{5cm}|}
\hline
Approach Category                  & Method                                               & Main Features                                                                                                          & Best Performance                                                            & Strengths                                                                                                                                                                                                                                   & Weaknesses                                                                                            \\ \hline
\multirow{4}{*}{Hand-crafted}      & \cite{koopman2018detection}         & Photo Response Non-Uniformity                                                                                          & High correlation for bonafide images than deepfakes in a self-build dataset & A simple feature that can be efficiently extracted                                                                                                                                                                                          & Evaluations have been conducted using a self-build dataset. Cannot handle unseen deepfake categories.  \\ \cline{2-6} 
                                   & \cite{kharbat2019image}             & Histogram of Gradient                                                                                                  & ACC=0.94 in UADFV dataset \cite{xie2020deepfake}                          & A simple feature that can be efficiently extracted                                                                                                                                                                                          & Can only handle face-swap deepfakes. Cannot handle unseen deepfake categories.                        \\ \cline{2-6} 
                                   & \cite{xia2022towards}               & texture difference from the colour channels                                                                            & AUC=0.99 in FF++ dataset \cite{rossler2019faceforensics++}                           & The framework is interpretable.                                                                                                                                                                                                             & Can only handle face-swap deepfakes. Cannot handle unseen deepfake categories.                        \\ \cline{2-6} 
                                   & \cite{wang2022ffr_fd}              & Speeded Up Robust Features (SURF), Scale-Invariant Feature Transform (SIFT), and Oriented Fast and Rotated Brief (ORB) & AUC=0.99 in DF-TIMIT(LQ) dataset \cite{korshunov2018deepfakes}                   & Can generalise to unseen datasets and different face deepfake generation methods                                                                                                                                                            & Can only handle face-swap deepfakes.                                                                  \\ \hline
\multirow{9}{*}{Artefacts}         & \cite{yang2019exposing}             & 3D head pose                                                                                                           & AUC=0.89 in UADFV dataset \cite{xie2020deepfake}                          & The framework is interpretable.                                                                                                                                                                                                             & Can only handle face-swap deepfakes. Cannot handle unseen deepfake categories.                        \\ \cline{2-6} 
                                   & \cite{xu2021deepfake}               & Gray-Level Co-occurrence Matrix                                                                                        & ACC=0.94 in DF-TIMIT(HQ) dataset \cite{korshunov2018deepfakes}                   & A simple feature that can be efficiently extracted                                                                                                                                                                                          & Can only handle face-swap deepfakes. Cannot generalise to unseen datasets.                            \\ \cline{2-6} 
                                   & \cite{kingra2022lbpnet}             & Local Binary Pattern (LBP)                                                                                             & AUC=0.99 in FF++ \cite{rossler2019faceforensics++} dataset                           & A simple feature that can be efficiently extracted. Can be used to detect both face swap and face reenactment categories. Can generalise to unseen datasets and different face deepfake generation methods. The framework is interpretable. & Cannot handle unseen deepfake categories.                                                             \\ \cline{2-6} 
                                   & \cite{agarwal2019protecting}        & head poses, facial landmarks, and expression                                                                           & AUC=0.96 in a self-build dataset                                            & The framework is interpretable. Can be used to detect both face swap and face reenactment categories.                                                                                                                                       & Evaluations have been conducted using a self-build dataset. Cannot handle unseen deepfake categories. \\ \cline{2-6} 
                                   & \cite{nguyen2020eyebrow}            & eyebrow                                                                                                                & AUC 0.88 in Celeb-DF dataset \cite{Celeb_DF_cvpr20}                        & Consistent performance using this feature as the input to different backbone feature extractors                                                                                                                                             & Can only handle face-swap deepfakes. Cannot handle unseen deepfake categories.                        \\ \cline{2-6} 
                                   & \cite{haliassos2021lips}            & mouth movement                                                                                                         & AUC=0.97 in DF1.0 \cite{jiang2020deeperforensics} dataset                          & Can be used to detect both face swap and face reenactment categories. Can generalise to unseen datasets and different face deepfake generation methods                                                                                      & The detection process is not interpretable                                                            \\ \cline{2-6} 
                                   & \cite{qi2020deeprhythm}             & Skin colour                                                                                                            & Acc=0.98 in FF++ \cite{rossler2019faceforensics++}                                   & Can be used to detect high-resolution face deepfakes                                                                                                                                                                                        & Can only handle face reenactment deepfakes. Cannot handle unseen deepfake categories.                 \\ \cline{2-6} 
                                   & \cite{fernandes2019predicting}      & oxygen concentration in the blood                                                                                      & Classification Loss of 0.0215 in a self-build dataset                       & Can be used to detect high-resolution face deepfakes                                                                                                                                                                                        & Can only handle face-swap deepfakes. Cannot handle unseen deepfake categories.                        \\ \cline{2-6} 
                                   & \cite{ciftci2020fakecatcher}        & remote PhotoPlethysmoGraphy                                                                                            & Acc=0.97 in UADFV \cite{xie2020deepfake} dataset                          & Can be used to detect both face swap and face reenactment categories.                                                                                                                                                                       & Cannot generalise to unseen datasets.                                                                 \\ \hline

\multirow{9}{*}{Deep Learning}     & \cite{afchar2018mesonet}            & Deep features extracted from a Capsule Network architecture                                                            & AUC=0.91 in a self-build dataset                                            & Can be used to detect both face swap and face reenactment categories.                                                                                                                                                                       & Cannot generalise to unseen datasets.                                                                 \\ \cline{2-6} 
                                   & \cite{kumar2020detecting}           & Deep features from ResNet-18                                                                                           & Acc=0.99 in FF++ dataset \cite{rossler2019faceforensics++}                           & A simplified framework for deepfake detection                                                                                                                                                                                               & Can only handle face reenactment deepfakes. Cannot handle unseen deepfake categories.                 \\ \cline{2-6} 
                                   & \cite{rana2020deepfakestack}        & Deep features from XceptionNet, MobileNet, ResNet101, InceptionV3, DensNet121, InceptionReseNetV2, and DenseNet169     & Acc=0.99 in a self-build dataset                                            & Can be used to detect both face swap and face reenactment categories.                                                                                                                                                                       & Cannot generalise to unseen datasets.                                                                 \\ \cline{2-6} 
                                   & \cite{guera2018deepfake}            & Deep features from a CNN + LSTM framework                                                                              & Acc=0.97 in a self-build dataset                                            & A simplified framework for deepfake detection in videos                                                                                                                                                                                     & Can only handle face-swap deepfakes. Cannot handle unseen deepfake categories.                        \\ \cline{2-6} 
                                   & \cite{nguyen2021learning}           & Deep features from a 3DCNN                                                                                             & Acc=0.99 in VidTIMID(HQ) \cite{sanderson2009multi} dataset                   & A simplified framework for deepfake detection in videos                                                                                                                                                                                     & Can only handle face-swap deepfakes. Cannot handle unseen deepfake categories.                        \\ \cline{2-6} 
                                   & \cite{wodajo2021deepfake}           & Deep features from a Convolutional Vision-Transformer                                                                  & Acc=0.93 in FF++ dataset \cite{rossler2019faceforensics++}                           & Can be used to detect both face swap and face reenactment categories.                                                                                                                                                                       & Cannot generalise to unseen datasets.                                                                 \\ \cline{2-6} 
                                   & \cite{mittal2020emotions}           & Deep learned emotion features extracted from audio and video                                                           & ACC=0.96 in DF-TIMIT(LQ) dataset \cite{korshunov2018deepfakes}                   & Can be used to detect both face swap and face reenactment categories. The framework is interpretable.                                                                                                                                       & Cannot generalise to unseen datasets.                                                                 \\ \cline{2-6} 
                                   & \cite{chugh2020not}                 & Deep features from 3D- ResNetand audio features from Mel-Frequency Cepstral Coefficients                               & ACC=0.97 in DF-TIMIT(LQ) dataset \cite{korshunov2018deepfakes}                   & The framework is interpretable.                                                                                                                                                                                                             & Can only handle face-swap deepfakes. Cannot handle unseen deepfake categories.                        \\ \cline{2-6} 
                                   & \cite{zhou2021joint}                & Deep Learned synchronisation features extracted from audio and video streams                                           & Acc=0.99 in FF++ dataset \cite{rossler2019faceforensics++}                           & Can be used to detect both face swap and face reenactment categories. Can generalise to unseen datasets. The framework is interpretable.                                                                                                    & Cannot handle unseen deepfake categories.                                                             \\ \hline
\multirow{4}{*}{Anomaly Detection} & \cite{khodabakhsh2020generalizable} & logarithmic probability of observing a particular pixel's intensity                                                    & Acc=0.98 in FF++ dataset \cite{rossler2019faceforensics++}                           & Can be used to detect both face swap and face reenactment categories. Can generalise to unseen datasets. The framework is interpretable.                                                                                                    & Cannot handle unseen deepfake categories.                                                             \\ \cline{2-6} 
                                   & \cite{wang2020exposing}             & local motion patterns                                                                                                  & Acc=0.98 in FF++ dataset \cite{rossler2019faceforensics++}                           & Can be used to detect both face swap and face reenactment categories. The framework is interpretable.                                                                                                                                       & Cannot generalise to unseen datasets.                                                                 \\ \cline{2-6} 
                                   & \cite{khalid2020oc}                 & Video reconstruction                                                                                                   & F1=0.98 in DFD dataset \cite{bhat2024dfda}                             & Can generalise to unseen datasets.                                                                                                                                                                                                          & Can only handle face-swap deepfakes. Cannot handle unseen deepfake categories.                        \\ \cline{2-6} 
                                   & \cite{cozzolino2023audio}           & Deep learned audio-visual features of authentic videos &  AUC=0.99 in DF-TIMIT dataset \cite{korshunov2018deepfakes}  & Can be used to detect both face swap and face reenactment categories.                                                                                                                                                                       & Cannot generalise to unseen datasets.                                                                 \\ \hline

\end{tabular}}
\label{tab:detection_summary}
\end{table*}



\section{Face deepfakes Biometric Evaluation}
In this section, we provide quantitative evaluations to demonstrate the ability of state-of-the-art face deepfake generation methods to fool advanced face recognition models.Face recognition systems are readily applied in numerous security-critical applications such as border control, authentication for banking apps, patient identification systems, and home automation. It should be noted that it is difficult for deepfakes to fool the physical biometric recognition systems such as systems used in border control. However, there exists evidence \footnote{https://edition.cnn.com/2024/02/04/asia/deepfake-cfo-scam-hong-kong-intl-hnk/index.html} that sophisticated deepfake technology can fool online authentication systems such as mobile-based personal authentication systems, as such, it is important to investigate the biometric implications of off-the-shelf face deepfake technology.

\subsection{Efficacy of face deepfakes to fool face biometrics systems}

\subsubsection{Evaluation Protocol}
In this evaluation, we follow a protocol similar to the one used in Sec. I.A. Specifically, we created face deepfakes using state-of-the-art face deepfakes generation methods: Wav2Lip \cite{chung2017out},  MCNet \cite{hong2023implicit}, First Order Motion Model \cite{siarohin2019first}, SimSwap \cite{chen2020simswap}, and FSGAN \cite{nirkin2022fsganv2}. From the training set of the Voxceleb2 \cite{nagrani2020voxceleb} dataset, we selected 36 subjects (with equal proportions of male and female subjects) and randomly selected 25 sample videos from each of those subjects. These 36 subjects were randomly paired as source and target faces and out of 25 sample videos that are available for each subject, 24 videos were selected to generate face deepfakes using both face swapping and face reenactment procedures. In the face swapping setting the facial components in the target face are replaced using source face features. In the face reenactment setting, the source video's expressions are replicated in the target video. 

State-of-the-art face recognition models, irse50 \cite{hu2018squeeze}, Facenet \cite{schroff2015facenet}, mobile face \cite{chen2018mobilefacenets}, ir152 \cite{deng2019arcface} and deep face \cite{serengil2024lightface} are used to verify the quality of the synthesised faces to fool the biometric systems in the face verification setting. Specifically, for face swap methods, the remaining video of the source subject out of the 25 video samples is used as the enrollment sample, and the generated 24 deepfake faces are verified biometrically against this sample. In contrast, in the face reenactment setting, the remaining video of the target subject is used for enrollment. 

\subsubsection{Evaluation Metric}
The Attack Success Rate (ASR) \cite{deb2020advfaces, zhong2020towards} is widely considered the evaluation metric to evaluate the effectiveness of attacks on face recognition methods. Let $F$ denote the backbone feature extractor of the face recognition model, $I_e$ denote the enrolled face image and $I_t$ denote the target image. Then, success rate, SR, can be defined as,

\begin{equation}
    SR = \frac{\sum_{i}^{N}1_{\tau}(cos[F(I^i_e), F(I^i_t)] > \tau)}{N} \times 100\%,
\end{equation}

where $N$ is the total number of image pairs that are being evaluated, $cos[X, Y]$ is a function that accepts two feature vectors and computes the cosine similarity between the vectors, and $\tau$ is a threshold that is being set based on the False Acceptance Rate (FAR) of the face recognition model.

However, more insights regarding the generated attacks under different circumstances can be generated by investigating the impact of different attack generation conditions on cosine similarity metric. Furthermore, different $\tau$ values should be utilised for different face recognition models to achieve a certain FAR. For instance, at 0.01 FAR for IR152 $\tau = 0.167$, IRSE50 $\tau =0.241$, MobileFace $\tau =0.302$ and Facenet $\tau =0.409$. Therefore, we also report the cosine similarity score between the vectors $I_v$ and $I_a$, which can be evaluated using,

\begin{equation}
    Similarity Score = \frac{F(I^i_e) . F(I^i_t)} {||F(I^i_e)|| \times ||F(I^i_t)||},
\end{equation}
where $.$ denotes the dot product of vectors.

\subsubsection{Results} 
Evaluation results of face deepfakes on face biometrics are presented in Tab. \ref{tab:face_evals}. The evaluations demonstrate the ability of deepfake methods such as Wav2Lip, SimSwap, and First-order-model to fool the biometric recognition systems, especially the lightweight systems such as MobileNet. This vulnerability is significantly concerning due to the vast utilisation of lightweight face verification methods for authentication in applications such as mobile device unlocking, app login and payment gateways, and in social media apps for photo tagging. 

\begin{table*}[htbp]
\caption{Evaluation of the efficacy of face deepfakes to thwart face biometrics systems. We evaluate both face swap methods (SimSwap, and FSGAN) and face reenactment methods (Wav2Lip, MCNet, and First Order Motion Model).}
\label{tab:face_evals}
\resizebox{\textwidth}{!}{%
\begin{tabular}{|c|cccc|cccc|cccc|cccc|cccc|}
\hline
\multirow{3}{*}{Model} & \multicolumn{4}{c|}{irse50}                                                                                            & \multicolumn{4}{c|}{Facenet}                                                                                           & \multicolumn{4}{c|}{Mobile Face}                                                                                       & \multicolumn{4}{c|}{ir152}                                                                                             & \multicolumn{4}{c|}{Deep Face}                                                                                        \\ \cline{2-21} 
                       & \multicolumn{3}{c|}{Success Rate}                                                  & \multirow{2}{*}{Similarity Score} & \multicolumn{3}{c|}{Success Rate}                                                  & \multirow{2}{*}{Similarity Score} & \multicolumn{3}{c|}{Success Rate}                                                  & \multirow{2}{*}{Similarity Score} & \multicolumn{3}{c|}{Success Rate}                                                  & \multirow{2}{*}{Similarity Score} & \multicolumn{3}{c|}{Success Rate}                                                 & \multirow{2}{*}{Similarity Score} \\ \cline{2-4} \cline{6-8} \cline{10-12} \cline{14-16} \cline{18-20}
                       & \multicolumn{1}{c|}{@01}  & \multicolumn{1}{c|}{@001} & \multicolumn{1}{c|}{@0001} &                                   & \multicolumn{1}{c|}{@01}  & \multicolumn{1}{c|}{@001} & \multicolumn{1}{c|}{@0001} &                                   & \multicolumn{1}{c|}{@01}  & \multicolumn{1}{c|}{@001} & \multicolumn{1}{c|}{@0001} &                                   & \multicolumn{1}{c|}{@01}  & \multicolumn{1}{c|}{@001} & \multicolumn{1}{c|}{@0001} &                                   & \multicolumn{1}{c|}{@01}  & \multicolumn{1}{c|}{@001} & \multicolumn{1}{c|}{@001} &                                   \\ \hline

SimSwap                & \multicolumn{1}{c|}{1.00} & \multicolumn{1}{c|}{1.00} & \multicolumn{1}{c|}{1.00}  & 0.96                              & \multicolumn{1}{c|}{0.97} & \multicolumn{1}{c|}{0.91} & \multicolumn{1}{c|}{0.85}  & 0.89                              & \multicolumn{1}{c|}{1.00} & \multicolumn{1}{c|}{1.00} & \multicolumn{1}{c|}{1.00}  & 0.97                              & \multicolumn{1}{c|}{1.00} & \multicolumn{1}{c|}{0.99} & \multicolumn{1}{c|}{0.98}  & 0.88                              & \multicolumn{1}{c|}{0.95} & \multicolumn{1}{c|}{0.89} & \multicolumn{1}{c|}{0.83} & 0.79                              \\ \hline
FSGAN                  & \multicolumn{1}{c|}{0.99} & \multicolumn{1}{c|}{0.95} & \multicolumn{1}{c|}{0.83}  & 0.43                              & \multicolumn{1}{c|}{0.58} & \multicolumn{1}{c|}{0.31} & \multicolumn{1}{c|}{0.08}  & 0.29                              & \multicolumn{1}{c|}{1.00} & \multicolumn{1}{c|}{0.99} & \multicolumn{1}{c|}{0.95}  & 0.56                              & \multicolumn{1}{c|}{0.63} & \multicolumn{1}{c|}{0.39} & \multicolumn{1}{c|}{0.23}  & 0.14                              & \multicolumn{1}{c|}{0.80} & \multicolumn{1}{c|}{0.24} & \multicolumn{1}{c|}{0.10} & 0.23                              \\ \hline
\hline

Wav2Lip                & \multicolumn{1}{c|}{1.00} & \multicolumn{1}{c|}{1.00} & \multicolumn{1}{c|}{1.00}  & 0.94                              & \multicolumn{1}{c|}{1.00} & \multicolumn{1}{c|}{1.00} & \multicolumn{1}{c|}{1.00}  & 0.95                              & \multicolumn{1}{c|}{1.00} & \multicolumn{1}{c|}{1.00} & \multicolumn{1}{c|}{1.00}  & 0.95                              & \multicolumn{1}{c|}{1.00} & \multicolumn{1}{c|}{1.00} & \multicolumn{1}{c|}{1.00}  & 0.855                             & \multicolumn{1}{c|}{1.00} & \multicolumn{1}{c|}{0.95} & \multicolumn{1}{c|}{0.87} & 0.79                              \\ \hline
MCNet                  & \multicolumn{1}{c|}{0.99} & \multicolumn{1}{c|}{0.95} & \multicolumn{1}{c|}{0.79}  & 0.39                              & \multicolumn{1}{c|}{0.26} & \multicolumn{1}{c|}{0.05} & \multicolumn{1}{c|}{0.01}  & 0.14                              & \multicolumn{1}{c|}{1.00} & \multicolumn{1}{c|}{0.92} & \multicolumn{1}{c|}{0.74}  & 0.43                              & \multicolumn{1}{c|}{0.19} & \multicolumn{1}{c|}{0.03} & \multicolumn{1}{c|}{0.02}  & 0.03                              & \multicolumn{1}{c|}{0.76} & \multicolumn{1}{c|}{0.01} & \multicolumn{1}{c|}{0.00} & 0.13                              \\ \hline
First Order Motion Model      & \multicolumn{1}{c|}{1.00} & \multicolumn{1}{c|}{1.00} & \multicolumn{1}{c|}{1.00}  & 0.98                              & \multicolumn{1}{c|}{1.00} & \multicolumn{1}{c|}{1.00} & \multicolumn{1}{c|}{1.00}  & 0.99                              & \multicolumn{1}{c|}{1.00} & \multicolumn{1}{c|}{1.00} & \multicolumn{1}{c|}{1.00}  & 0.99                              & \multicolumn{1}{c|}{1.00} & \multicolumn{1}{c|}{1.00} & \multicolumn{1}{c|}{1.00}  & 0.94                              & \multicolumn{1}{c|}{1.00} & \multicolumn{1}{c|}{0.95} & \multicolumn{1}{c|}{0.92} & 0.88                              \\ \hline

\end{tabular}}
\end{table*}

\section{Future research directions}
In this section, we outline the limitations of existing deepfake generation and detection techniques as well as various open research questions, and highlight future research directions.

\subsection{Universal deepfake detection methods}
Deepfake generation technology evolves over time and it is practically impossible to consider every possible generation type, that could be introduced in the future, into consideration. As such, developing universal deepfakes detection technology that could withstand not only the current deepfakes but also the advances that the deepfake generation methods would attain in the future is virtually impossible. 

Furthermore, the variations of domains and contexts make the generalisation of the deepfake detectors challenging. For instance, a universal deepfake detector should be generalisable to the low-quality media shared on social media, as well as high-quality deepfakes produced in the entertainment industry. Furthermore, it should be generalisable to different backgrounds, modalities, etc. These diverse real-world settings make it a significant challenge. 

We observe a further hindrance to achieving universal detection due to the limited data availability for model training. Specifically, the supervised training of a universal deepfake detector will require a myriad of training data that span various deepfake techniques, modalities, quality levels, resolutions, and content types, which are not readily available. Furthermore, the computational complexity of training a large-scale, complex deep learning model could impede the design of a universal detector. 

Future research efforts could be directed to address these challenges. One possible avenue for exploration is the use of biometric features for the generation of subject-specific deepfake detection. In addition to traditional biometric features, complementary behavioral biometric and linguistic features could be used to supplement the model learning. Moreover traditional machine learning tools and physics or biologically inspired architectures could be leveraged for the design of universal deepfakes detection methodologies.

\subsection{Recovering the true identity}

Legal and law enforcement are not the only applications that require revealing of the true identity of a person in a deepfake video. Revealing of the true identity helps mitigate the damage caused to a person's reputation and could stop the spread of malicious media. Furthermore, the systems that can validate media for the presence of deepfakes and recover the true identity of a person (if it is identified as deepfake media) could be embedded in social media platforms. Such actions could help build trust among the general public about the content shared on those platforms. 

To the best of our knowledge the only work that is capable of restoring the true identity of a subject in a deepfake video is proposed in the cyber immune system framework proposed in \cite{chang2023cyber}. However, there exist several limitations of this work and numerous future research directions. For instance, the authors of \cite{chang2023cyber} have pointed out that the recovery of the true identity is impossible in their framework if there exist major pose changes during the face reenactment. Furthermore, this model works only for the front-facing face images and produces inferior results under poor illumination conditions. 

Future research efforts could also be made towards recovering the true identity in multimodal deepfake videos. The existence of two or more modalities can be seen as mediums to embed and extract complementary identity features related to the true identity. Furthermore, we observe the possibilities of extending the framework in \cite{chang2023cyber} to the recovery of true identity in full-body deepfakes. Posture and body movements, hand gestures, facial expressions, and micro expressions all carry informative cues to recover the true identity of a person in a deepfake video, therefore, future efforts could be directed to investigate how those physical and behavioral characteristics can be incorporated into the cyber vaccine framework of \cite{chang2023cyber}.

\subsection{Explainable deepfakes detection methods}

Explainability is a paramount characteristic of any machine learning algorithm. As free online deepfake detection tools such as Resemble.ai deepfake detector \footnote{https://www.resemble.ai/free-deepfake-detector/}, Deepware deepfake scanner \footnote{https://scanner.deepware.ai/}, and the deepfake detector \footnote{https://deepfakedetector.ai/} are increasingly becoming popular among general public explainability has become a curial characteristic to maintain the trust and transparency of the detection results. The generated explanations will build trust and confidence among the users and the general public regarding the reliability and fairness of the detection process. Furthermore, they can be leveraged to train the users how to detect deepfakes without using such tools and elevate their literacy. Such education, which could be given to employees, would have a significant impact on strengthening protective measures and mitigating the risks of cyber espionage through impersonation.

Recently a few research efforts \cite{xu2022supervised, ishrak2022explainable, mathews2023explainable} have been made to preserve the explainability of the deepfake detection methods. However, more investigations along these lines are encouraged such that explainable detection of deepfakes can be achieved in all the categories of deepfake generation algorithms, including, voice, face, full-body, and multimodal deepfakes.

\subsection{Introducing standard evaluation protocols} 

We observe inconsistencies in the evaluation metrics used in both deepfake generation and detection literature. For instance, in face deepfake generation several studies have used metrics such as average Euclidean distance between the feature vectors for real and generated faces, peak signal-to-noise ratio, pixel-wise distances with respect to facial landmarks, etc. In contrast, some studies \cite{wu2018reenactgan} utilise human surveys to compare the quality of the media that have been generated by different deepfake generation methods. Therefore a unified evaluation protocol is needed to properly assess and validate the effectiveness of a particular method. 

Similarly in face deepfake detection literature precision, recall, and F1-Score have been the most commonly used metrics for comparison. However, some studies have reported their performance using AUC and Error-Rate metrics which makes comparison across different state-of-the-art methods infeasible. 

\textbf{Accuracy:} measures the proportion of the samples in the test set that have been correctly classified by the detection algorithm.  \textbf{Precision: } measure the ratio of the correctly identified positive instances (eg. fake samples) out of the total positive samples that the model classified as positive which could be denoted as, \[\text{Precision} = \frac{\text{True Positives}}{\text{True Positives} + \text{False Positives}}.\] Higher precision denotes that the model is making a lesser number of false positive identifications. On the other hand \textbf{Recall} determines the model's ability to correctly identify a positive data sample. It is calculated as the proportion of true positives out of all the positive samples in the test dataset and could be written as, \[\text{Recall: } = \frac{\text{True Positives}}{\text{True Positives} + \text{False Negatives}}.\] Higher recall is a vital attribute for deepfake detection algorithms as misclassification of a true positive sample in a real-world application could be costly. \textbf{F1-Score: } could be calculated as \[\text{F1-Score} = 2 \times \frac{\text{Precision} \times \text{Recall}}{\text{Precision} + \text{Recall}},\] which is the harmonic mean of both precision and recall metrics.  \textbf{AUC: } measures the ability of the model to distinguish the fake samples from the real samples and is calculated by integrating the ROC Curve. \textbf{Error Rate (EER):} represents the proportion of incorrectly classified samples out of the total number of samples in the test dataset. It could also be calculated as \[\text{EER} = 1 - \text{Accuracy}.\]

Standard evaluation metrics are essential to promote rigorous, transparent, and meaningful comparisons across different models proposed in the literature. They remove the subjectivity or the bias from the evaluation process which results in fair comparisons. Furthermore, standard metrics promote the independent reproducibility of the results so that they can be independently validated. Furthermore, for decision making such as the deployment of a certain model in industrial applications, the stakeholders need to compare metrics of different state-of-the-art methods. As such, it is important to maintain consistency in evaluation methods.

\subsection{Design of a regulatory framework for the governance of deepfake research}
The researchers should also investigate and establish a regulatory framework for the governance of deepfake research. They should proactively collaborate with relevant stakeholders, including, governments, law enforcement authorities, and the general public to ensure that future research into deepfakes is conducted in a transparent and accountable manner, and addresses ethical and privacy concerns. Most importantly a governance framework will establish ethical guidelines and standards for the creation and distribution of deepfakes, and the public share of the implementations. As such, this framework has the potential of minimising the negative societal implications of deepfakes due to misuse. Furthermore, the establishment of such a regulatory framework would promote collaboration among researchers, industry stakeholders, policymakers, and advocacy groups which could foster innovation in a responsible manner. The process of designing governance protocols could also be used to enhance public education and awareness regarding both the merits and negative implications of deepfake technology which will help uphold societal trust in AI. 

In addition to a regulatory framework the researchers could look into a philosophical and moral framework to establish a set of principles, values and guidelines to make decision regarding conducting research in the domain of deepfakes. When establishing this framework the board societal impact of deepfakes should be taken into consideration. Fairness, integrity, and compassion will be among the foundational values upon which this framework will be built and researchers from different domains, including, legal, philosophical and technical research areas should collaborate when establishing the principles and best practices that encompass this framework.

\bibliographystyle{IEEEtran}
\bibliography{bib/egdb, bib/deepfake, bib/fullbody}

\end{document}